\documentclass[journal]{IEEEtran}
\usepackage{amsmath,amsfonts}
\usepackage{algorithmic}
\usepackage{algorithm}
\usepackage{array}
\usepackage[caption=false,font=normalsize,labelfont=sf,textfont=sf]{subfig}
\usepackage{textcomp}
\usepackage{stfloats}
\usepackage{url}
\usepackage{verbatim}
\usepackage{graphicx}
\usepackage{cite}
\usepackage{multirow} 
\usepackage{xcolor}

\usepackage{amsmath}
\usepackage{enumitem}
\usepackage{hyperref}

\newcommand{\model}{Mesh2SSM++}
\newcommand{\X}{\mathbf{X}}
\newcommand{\V}{\mathbf{V}}
\newcommand{\C}{\mathbf{C}}
\newcommand{\z}{\mathbf{z}}

\newcommand{\ve}{\mathbf{v}}
\newcommand{\ci}{\mathbf{c}}

\newcommand{\set}[1]{\mathcal{#1}}
\newcommand{\realdim}{\mathbb{R}}

\begin{document}

\title{\model: A Probabilistic Framework for Unsupervised Learning of Statistical Shape Model of Anatomies from Surface Meshes}

\author{Krithika Iyer, \and Mokshagna Sai Teja Karanam,
\and Shireen Elhabian\\
{\{krithika.iyer@, u1418261@, shireen@sci\}.utah.edu}
\thanks{K. Iyer, M. Karanam, and S. Elhabian are with the Scientific Computing and Imaging Institute and the Kahlert School of Computing, University of Utah, Utah, USA}}

\markboth{Journal of \LaTeX\ Class Files,~Vol.~14, No.~8, August~2021}%
{Shell \MakeLowercase{\textit{et al.}}: A Sample Article Using IEEEtran.cls for IEEE Journals}


\maketitle

\begin{abstract}
Anatomy evaluation is crucial for understanding the physiological state, diagnosing abnormalities, and guiding medical interventions. Statistical shape modeling (SSM) is vital in this process, particularly in medical image analysis and computational anatomy. By enabling the extraction of quantitative morphological shape descriptors from medical imaging data such as MRI and CT scans, SSM provides comprehensive descriptions of anatomical variations within a population. However, the effectiveness of SSM in anatomy evaluation hinges on the quality and robustness of the shape models, which face challenges due to substantial nonlinear variability in human anatomy. While deep learning techniques show promise in addressing these challenges by learning complex nonlinear representations of shapes, existing models still have limitations and often require pre-established shape models for training. To overcome these issues, we propose \model, a novel approach that learns to estimate correspondences from meshes in an unsupervised manner. This method leverages unsupervised, permutation-invariant representation learning to estimate how to deform a template point cloud into subject-specific meshes, forming a correspondence-based shape model. Additionally, our probabilistic formulation allows learning a population-specific template, reducing potential biases associated with template selection. A key feature of \model~is its ability to quantify aleatoric uncertainty, which captures inherent data variability and is essential for ensuring reliable model predictions and robust decision-making in clinical tasks, especially under challenging imaging conditions. Through extensive validation across diverse anatomies, evaluation metrics, and downstream tasks, we demonstrate that \model~outperforms existing methods. Its ability to operate directly on meshes, combined with computational efficiency and interpretability through its probabilistic framework, makes it an attractive alternative to traditional and deep learning-based SSM approaches. \href{https://github.com/iyerkrithika21/Mesh2SSMJournal}{Github: https://github.com/iyerkrithika21/Mesh2SSMJournal}
\end{abstract}


\section{Introduction} \label{introduction}
Anatomy evaluation is the systematic quantitative assessment of the human body's form and function, crucial for comprehending the medical condition, diagnosing abnormalities, and guiding medical interventions \cite{singh2020evaluation,dai2020statistical,zhang20243dcmm,quiceno2024statistical}. Statistical shape modeling (SSM) is essential for anatomy evaluation in medical image analysis and computational anatomy. SSM facilitates the discovery of quantitative morphological shape descriptors that comprehensively describe anatomical variations within the context of the population, aiding in diagnosis \cite{khan2022machine,schaufelberger2022radiation}, pathology detection \cite{peiffer2022statistical,sophocleous2022feasibility}, treatment planning \cite{vicory2022statistical}, and enhancing early intervention strategies \cite{merle2019high,mulder2022dynamic,okegbile2022human}.

\begin{figure}
    \centering
    \includegraphics[width=\linewidth]{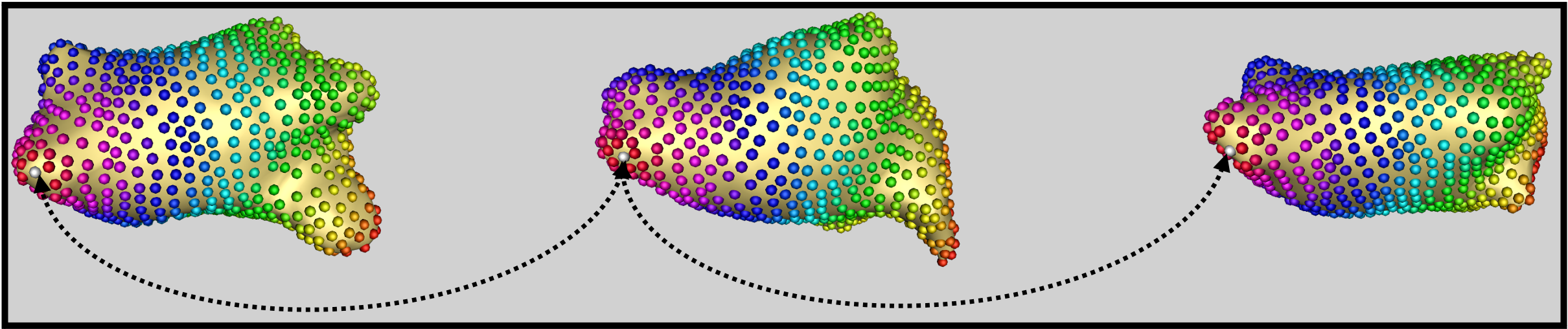}
    \caption{Correspondences are sets of ordered points on different shapes representing the same anatomical or geometric feature, thereby establishing a consistent relationship between the shapes. The white highlighted points represent predicted correspondences by \model~in the right superior pulmonary vein (RSPV) antrum region of the left atrium, consistently located across all shapes. Matching colors across samples indicate additional corresponding points. }
    \label{fig:correspondences_example}
    \vspace{-8mm}
\end{figure}

SSM methods have traditionally relied on two primary approaches for representing anatomical structures: \textit{implicit} representations (e.g., deformation fields \cite{durrleman2014morphometry} and level set methods \cite{samson2000level}) and \textit{explicit} representations (e.g., Point Distribution Models, or PDMs, which use ordered set of landmarks or correspondence points). Figure~\ref{fig:correspondences_example} shows examples of PDM for three different anatomies. The points highlighted in white represent these correspondence points, and the matching colors across samples indicate the established correspondences. Traditional SSM methods establish correspondences via (a) pairwise methods that map a shape subject to a pre-defined atlas or template via fixed geometrical bases (SPHARM-PDM \cite{styner2006framework}) or diffeomorphic metric mapping (Deformetrica \cite{durrleman2009statistical}), (b) group-wise approaches that establish the correspondences by considering the variability of the entire shape cohort (Particle-based Shape Modeling \cite{cates2007shape,cates2008particle}). 
However, the effectiveness of SSM in anatomy evaluation critically depends on the quality and robustness of the models, which must accurately capture real-world anatomical variability while handling noise and incomplete data \cite{cerrolaza2019computational}.

Despite their utility, traditional SSM approaches face several significant limitations. They often rely on computationally expensive optimization frameworks and manually tuned parameters, making them non-scalable and inefficient for handling large datasets. Their dependence on linearity assumptions further restricts their ability to model non-linear anatomical variations, leading to limited generalization. Moreover, traditional methods require full recomputation of the model for new samples, rendering them impractical for real-time or large-scale clinical applications. Deep learning-based approaches for SSM have emerged as a promising avenue for streamlining SSM. Deep learning models can learn complex non-linear representations of the shapes, which can be used to construct shape models. Moreover, they can efficiently perform inference on new samples without computation overhead or re-optimization. Unsupervised methods \cite{iyer2023mesh2ssm,adams2023point2ssm,ludke2022landmark,bhalodia2024deepssm,bhalodia2018deepssm,el2024universal,kalaie2024end} have demonstrated the ability to estimate population-level correspondences from meshes, point clouds, and images while eliminating the need to repeat the shape modeling pipeline for unseen samples and maintaining computational efficiency at inference. Specifically, Mesh2SSM \cite{iyer2023mesh2ssm} is an unsupervised deep learning approach that generates statistical shape models directly from surface meshes by deforming a template point cloud to subject-specific meshes. It eliminates the bias arising from template selection by incorporating a VAE \cite{kingma2019introduction} on the learned correspondences that help learn the underlying manifold and enable sampling a population-informed template. 

However, Mesh2SSM has certain limitations. The incorporation of VAE complicates the training process as the end-to-end training can be tricky as VAE requires careful parameter tuning to avoid posterior collapse. Mesh2SSM uses Chamfer distance to train the network such that the predicted correspondences faithfully represent the underlying shape, but this does not guarantee that the predicted particles will lie on the surface of the mesh. Medical datasets pose additional challenges, including data limitations and the need for uncertainty quantification to avoid overconfident predictions and provide clinicians with insights into model limitations. Real-world medical data often contains artifacts like noise and blurred edges, creating ambiguity in estimating correspondence positions, particularly near sharp edges or in noisy regions. This ambiguity necessitates accounting for aleatoric uncertainty, stemming from inherent data variability. Addressing this is essential for reliable outcomes in medical applications where precise modeling is paramount. Mesh2SSM does not account for these scenarios or provide any insights into the confidence of the model predictions. 

Therefore, we introduce \model, an unsupervised learning framework for constructing correspondence-based probabilistic statistical shape models directly from surface meshes. Building upon its predecessor, Mesh2SSM \cite{iyer2023mesh2ssm}, the proposed \model~introduces several key improvements to aid in effective anatomy evaluation. Our contributions are summarized as follows: 
\begin{enumerate}
        \item \textbf{Flow-enhanced latent space for probabilistic shape modeling: }Introduction of a continuous normalizing flow \cite{rezende2015variational,dinh2016density} in the latent space of the mesh autoencoder. This enhancement streamlines the model architecture while endowing the framework with powerful probabilistic capabilities. The normalizing flow enables:
        \begin{itemize}
            \item High-quality sample generation, owing to the invertibility of the transformations of the flows, facilitates a more accurate representation of anatomical variability.
            \item Streamlined data-informed template updates with simplified end-to-end training of a probabilistic network. 
            \item Efficient aleatoric uncertainty estimation for the correspondence prediction task, providing crucial reliability measures in medical applications.
        \end{itemize}
        \item \textbf{Mesh-constrained correspondence prediction:} We introduce a new loss function to encourage predicted correspondences to lie on the mesh surface, thereby improving the quality and anatomical accuracy of the correspondences.
        \item \textbf{Self-supervised training to increase robustness to noisy data:} We introduce vertex masking as an additional form of data augmentation, enhancing the model's robustness to noisy input and improving generalization.
        \item \textbf{Comprehensive evaluation:} We conduct comprehensive experimentation on multiple anatomical datasets, employing a wide range of evaluation metrics to assess model performance thoroughly. Our rigorous benchmarking against state-of-the-art (SOTA) mesh-based models demonstrates the effectiveness and advantages of \model~in medical shape analysis. We also show the utility of \model~by assessing its performance for two downstream tasks. 
\end{enumerate}

Overall, \model~enhances shape descriptor extraction, improves efficiency in correspondence generation, and extends SSM applicability to diverse anatomical structures. These advancements make \model~a robust, scalable solution for clinical applications requiring accurate and interpretable shape models.
\section{Related Work}

The increasing complexity of anatomical shapes and the volume of data in modern medical imaging have rendered manual landmarks/correspondence annotation impractical. This has led to a reliance on computational methods to establish anatomical correspondences. Methods to establish correspondences broadly include (a) registration-based landmark estimation and (b) parametric and non-parametric correspondence optimization. Registration-based methods involve manually annotating landmarks on a reference shape and warping these landmarks to other shapes using registration techniques \cite{paulsen2002building,heitz2005statistical,mcinerney1996deformable}. Parametric methods use fixed geometrical bases to derive correspondences \cite{styner2006framework}, but they often struggle with complex anatomical shapes due to limited flexibility. Non-parametric approaches establish correspondences based on cohort-level variability by optimizing objective functions \cite{cates2008particle,cates2017shapeworks,dai2020statistical}. A notable family of tools employs entropy-based objectives to establish correspondences \cite{oguz2016entropy,cates2007shape,cates2008particle,cates2017shapeworks} across a cohort of shapes while ensuring faithful representation of each shape. Particle-based Shape Modeling (PSM) has emerged as a state-of-the-art method in this category.

However, conventional methods face multiple challenges. Recent advancements in deep learning have significantly improved SSM by enabling the efficient learning of non-linear shape representations. Mesh-based and point cloud-based deep learning methods have emerged as promising alternatives due to their ability to model the structured representation of 3D shapes. Recent advancements in point cloud-based SSM, such as Point2SSM \cite{adams2023point2ssm}, Point2SSM++ \cite{adams2022images}, and other point cloud-based approaches \cite{lang2021dpc,wang2019dynamic,chen2020unsupervised} have demonstrated the potential of using raw point clouds for constructing shape models. These methods offer advantages in terms of reduced computational burden and relaxed input requirements. However, they lack the crucial connectivity information inherent in mesh-based approaches. Mesh-based SSM methods leverage this connectivity to better capture local geometric relationships and preserve fine anatomical details, which is particularly important for accurately modeling complex structures. Therefore, despite the progress in point cloud-based shape modeling, mesh-based approaches remain essential for comprehensive and precise statistical shape modeling in medical applications.

Early mesh-based networks, such as ShapeNet \cite{chang2015shapenet} and 3D-R2N2 \cite{choy20163d}, introduced volumetric grids and multi-view projections for mesh classification and vertex segmentation but faced challenges with high computational costs and resolution limitations. Mesh-specific neural networks, such as GCNN \cite{zhang2019graph,kipf2016semi} and MeshCNN \cite{hanocka2019meshcnn}, extended convolution operations to irregular graph-like structures, enabling effective feature extraction while preserving geometric details. Spectral methods like ChebNet \cite{defferrard2016convolutional} provided further advancements by operating in the spectral domain, and dynamic graph convolution networks (DGCNN) \cite{wang2019dynamic} introduced dynamic edge connections to improve the capture of local geometric relationships. Although ChebNet excels at modeling global structures, DGCNN spatial adaptability makes it particularly suitable for medical SSM applications that require localized and flexible shape modeling. 

Recent methods have addressed the challenges of shape vertex matching using unsupervised learning approaches. ShapeFlow \cite{jiang2020shapeflow} learns a deformation space for 3D shapes with large intra-class variations, using neural networks to parameterize continuous flow fields between a pair of meshes. FlowSSM \cite{ludke2022landmark} extended the idea to model population variation using neural networks to parameterize the deformations field between a cohort of shapes and a template in a low dimensional latent space and rely on an encoder-free setup. However, FlowSSM exhibits sensitivity to template selection and lacks an encoder, requiring re-optimizing latent representations for unseen shapes, increasing computational overhead \cite{iyer2023mesh2ssm}. Atlas-R-ASMG \cite{kalaie2024end} is an end-to-end deep learning generative framework for refinable shape matching and generation using 3D surface mesh data. The framework jointly learns high-quality refinable shape matching and generation while constructing a population-derived atlas model, enabling the generation of diverse and realistic anatomical shapes for virtual populations. However, Atlas-R-ASMG still relies on vertex-level correspondence by employing an attention mechanism in latent space to measure the similarity between local vertex embeddings between the atlas and the shapes. This approach, while effective, may limit the framework's flexibility in handling highly variable anatomical structures or shapes with significant topological differences. Addressing these limitations, Mesh2SSM \cite{iyer2023mesh2ssm} introduces a novel approach by replacing the encoder-free setup of FlowSSM with geodesic features and an EdgeConv-based \cite{wang2019dynamic} mesh autoencoder. This method overcomes the issues its predecessors face by eliminating reliance on vertex-level correspondences. Instead, Mesh2SSM produces a correspondence model with a fixed number of landmarks determined by the initial template. This is typically smaller than the total number of vertices, resulting in a more compact representation. Mesh2SSM also introduces a variational autoencoder \cite{kingma2019introduction} for learning a data-driven template from the predicted correspondences. However, Mesh2SSM cannot ensure predicted correspondences lie on the surface of the mesh, does not include aleatoric uncertainty quantification for predictions, and can have instabilities in training due to the VAE used for analysis. 

Our work builds on recent progress, particularly Mesh2SSM's strengths while addressing its limitations. We propose a method that introduces a probabilistic framework with an end-to-end training strategy, improved sample generation, aleatoric uncertainty estimation, and enhanced correspondence quality across variable anatomical structures.
\section{Methods}\label{methods}
\begin{figure}
    \centering
    \includegraphics[width=1\linewidth]{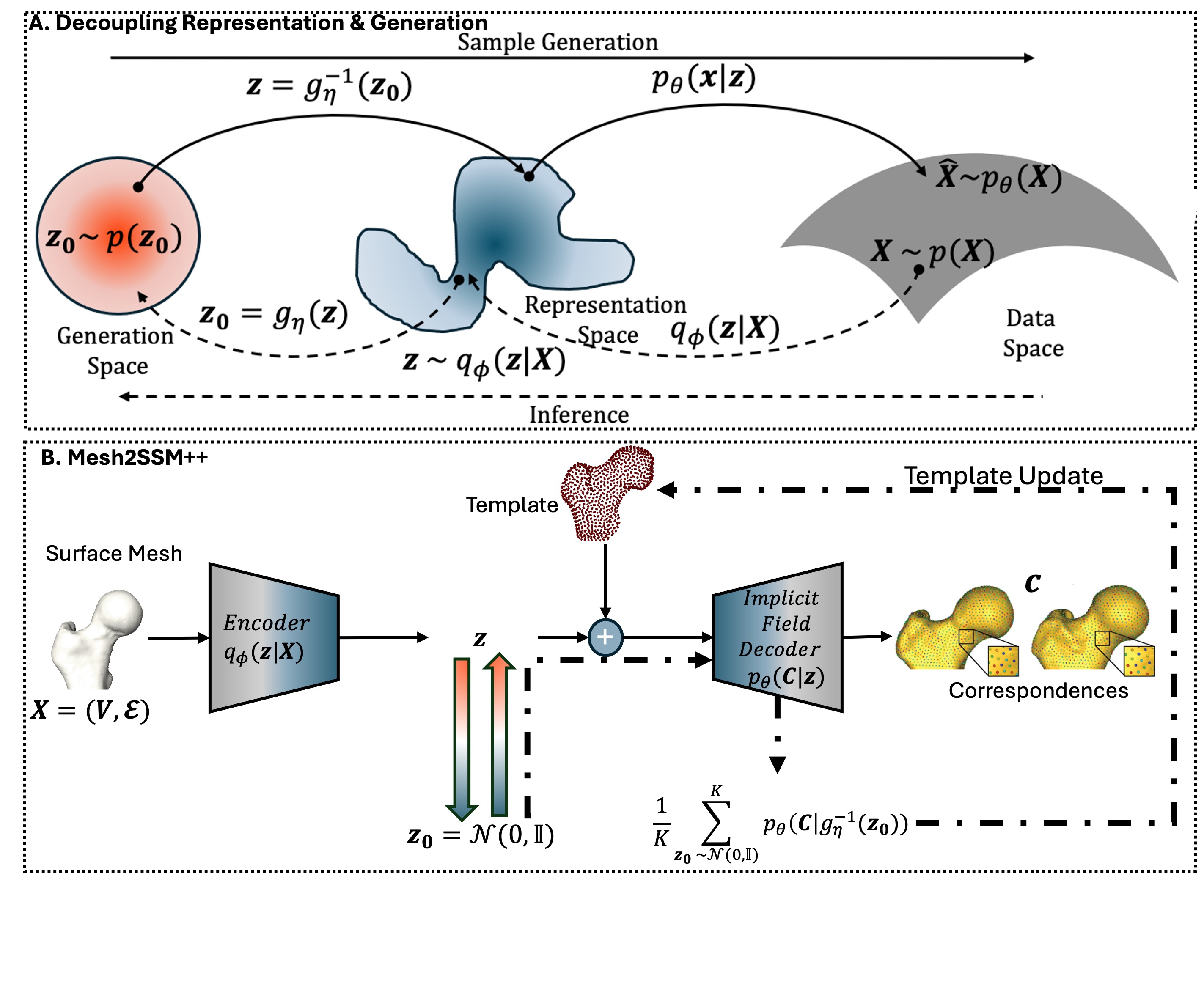}
    \vspace{-15mm}
    \caption{\textbf{Overview of \model~Framework}: (A) The generative model leverages a decoupled representation and generation process. The latent variable \(\z\) is mapped from the generation space \(\z_0 \sim p(\z_0) \) to the representation space through the invertible mapping \( g_\eta^{-1}(\z_0) \), while \(\set{X} \sim p_\theta(\set{X}|\z) \) represents data sampled in the data space. Inference is performed via \( q_\phi(\z|\set{X}) \), which maps the input mesh \(\set{X}\) to its latent representation \(\z\). (B) The \model pipeline. Input surface meshes \( \set{X}_i = (\set{V}_i, \set{E}_n) \) are processed by the encoder \( q_\phi(\z|\set{X}) \) to produce latent representations \(\z \). These are combined with prior samples \(\z_0 \sim \mathcal{N}(0, I) \) for probabilistic sampling. Based on the latent representation \(\z\), the implicit field decoder \( p_\theta(\set{X}|\z)\) learns how to deform the template point cloud into \(\C\) such that it matches the input shape surface while establishing correspondence by deforming the same template for every input mesh.}
    \vspace{-3mm}
    \label{fig:model_arch}
\end{figure}
This section provides a brief overview of Mesh2SSM and the proposed enhancements provided for \model. 

\subsection{Notation}\label{notations}
Consider a collection of \(N\) surface meshes, denoted as \(\set{X} = \{\X_1, \X_2, \dots, \X_N\}\). Each mesh \(\X_n\) is defined by \(K\) vertices \(\V_n = \{\ve_{n}^{(k)}\}_{k=1}^K \) where \(\ve_n^{(k)} \in \realdim^3\), and edge connectivity \(\set{E}_n\). Mesh2SSM aims to identify \(M\) correspondence points \(\C_n = \{\ci_{n}^{(m)}\}_{m=1}^M \) where \(\ci_{n}^{(m)} \in \realdim^3\) that accurately represent the anatomy of \(\X_n\) while maintaining anatomical consistency across \(\set{X}\). 
The number of correspondences \(M\) is typically less than the number of vertices \(K\) because it allows for a more compact and efficient representation of shape variability. By focusing on key anatomical landmarks or regions of interest, fewer correspondences can effectively capture the essential shape features while reducing computational complexity and improving statistical robustness across different mesh resolutions.

\subsection{Mesh2SSM}\label{mesh2ssm}

\subsubsection{Architecture}
Mesh2SSM is an unsupervised deep learning framework for generating statistical shape models directly from surface meshes. The method comprises two primary components: a \textit{correspondence generation module} and an \textit{analysis module}, working in tandem to create accurate and consistent shape representations.
\begin{itemize}
    \item \textit{Correspondence generation module} is designed to establish anatomically consistent correspondences across a set of surface meshes \(\set{X}\). This module comprises two primary networks: a Mesh Autoencoder (M-AE) and an Implicit Field Decoder (IM-NET). The M-AE, based on the Dynamic Graph CNN (DGCNN) architecture \cite{wang2019dynamic}, employs EdgeConv blocks to capture local, permutation-invariant geometric features of the input mesh. The first EdgeConv block utilizes geodesic distance on the mesh surface for feature calculation, enhancing its ability to capture intrinsic surface properties. The M-AE learns a low-dimensional representation \(\z_n \in \realdim^L\) for each mesh \(\X_n\), effectively encoding the geometric characteristics of the mesh. The learned representation \(\z_n\) is input to the IM-NET \cite{chen2019net}, a neural network designed for shape deformation. IM-NET employs a global shape descriptor \(\z_n\) to learn how to deform a template point cloud, adjusting each point's location individually to match the shape of each input sample. This process is applied consistently across all samples using the same initial point cloud, thereby implicitly establishing point-to-point correspondences across all meshes in the dataset. The module's optimization is guided by a loss function that combines point-set Chamfer distance (between predicted correspondences and ground truth vertices) and vertex reconstruction loss (see Eq~\ref{mae_equation}). This comprehensive approach enables efficient parameterization of surface meshes. Here, \(\alpha\) and \(\gamma\) are hyperparameters and \(\hat{\V}_n\) is the reconstructed vertex locations.  
    \begin{equation}\label{mae_equation}
    \begin{aligned}
    \mathcal{L}_{C} &= \sum_{n=1}^N \Big[ \mathcal{L}_{L_2 Chamfer}(\V_n,\C_n) 
    + \\ 
    &\quad  \alpha \mathcal{L}_{L_1 Chamfer}(\V_n,\C_n) 
    + \gamma \mathcal{L}_{MSE}(\V_n, \hat{\V}_n) \Big]
    \end{aligned}
    \end{equation}

    \item \textit{Analysis module} incorporates a Shape Variation Autoencoder (SP-VAE) that operates directly on predicted correspondences to capture non-linear shape variations from the learned correspondences. This VAE \cite{kingma2019introduction} maps the correspondence points to a latent space and reconstructs them, allowing for the estimation of mean shape and shape variations. It generates multiple samples from the latent space, which are then averaged to create a data-informed template. This template is periodically fed back into the correspondence generation module during training, refining the model's understanding of the underlying shape distribution. SP-VAE maintains the exact ordering of correspondences at input and output, ensuring consistency, and is parameterized by an encoder \(\phi\), decoder \(\theta\), and the prior \(p(\z) = \mathcal{N} (\mathbf{0}, \mathbf{I})\). The SP-VAE is trained using the following loss function:
    \begin{equation}\label{vae_equation}
    \begin{aligned}
    \mathcal{L(\theta,\phi)} &= -\mathbb{E}_{q_{\phi} (\z_n|\C_n)} \left[\operatorname{log} p_\theta(\C_n|\z_n)\right] + \\
    &\quad \operatorname{KL}[q_{\phi}(\z_n|\C_n) || p(\z_n)]      
    \end{aligned}
    \end{equation}
\end{itemize}
\subsubsection{Training}
The training process begins with a burn-in phase that prioritizes training the correspondence generation module (Eq.~\ref{mae_equation}), followed by alternating optimization of correspondence (Eq.~\ref{mae_equation}) and analysis (Eq.~\ref{vae_equation}) modules. To create a data-informed template, 500 instances from the prior distribution \(p(\z)\) are sampled and then decoded by the SP-VAE to reconstruct the correspondence point set. The mean template is derived from the average of these reconstructed samples and is subsequently used in successive epochs as the template point cloud input to the IM-NET. For inference with unseen meshes, they are passed through the mesh encoder and IM-NET to predict correspondences.

\subsection{\model: Mesh to Probabilistic Shape Model}\label{mesh2ssm_plus}

Building upon the foundation of Mesh2SSM, we propose \model, a method that introduces significant improvements to address existing limitations and enhance performance in SSM. The key advancements are as follows:

\begin{enumerate}

\item \textit{Normalizing Flow in Latent Space:}  
To enhance the probabilistic modeling of shape distributions, we replace the SP-VAE in our architecture with a normalizing flow (NF) framework \cite{rezende2015variational, dinh2016density}. This approach addresses limitations of VAEs, such as mode collapse and complex hyperparameter tuning \cite{alemi2018fixing}, while reducing the overall network complexity and enabling efficient end-to-end training.  

As shown in Figure~\ref{fig:model_arch}.A, in the proposed inclusion of NF divides the latent space into two complementary components:  

1. Representation Space (\(\z\)): Encodes global, semantically meaningful features of the input surface, derived using the DGCNN encoder.  

2. Sampling Space (\(\z_0\)): Defines a simpler latent space with a standard Gaussian prior, \(p(\z_0) = \mathcal{N}(\mathbf{0}, \mathbf{I})\), enabling effective probabilistic sampling.  

A bijective mapping \(g_\eta\) is introduced to transform between these spaces. Specifically,  
\begin{equation}\label{cnf_transformation}
\begin{aligned}
p_{\eta}(\z) &= p(\z_0) \left| \frac{\partial \z_0}{\partial \z} \right| = p(\z_0)\left| \frac{\partial g_{\eta}(\z)}{\partial \z} \right| 
\end{aligned}
\end{equation}
where \(\eta\) represents the network parameters of NF. The NF transforms the Gaussian prior \(p(\z_0)\) into a more expressive latent distribution \(p_\eta(\z)\) via the change-of-variable formula:  
\begin{equation}\label{cnf_transformation_log}
\begin{aligned}
\operatorname{log}p_\eta(\z) = \operatorname{log}p(\z_0) + \operatorname{log}\left|\det \frac{\partial g_\eta(\z)}{\partial \z} \right|,
\end{aligned}
\end{equation}
where the second term, the log-determinant of the Jacobian of \(g_\eta\), is computed efficiently using continuous normalizing flows (CNFs) \cite{chen2023learning}.  
\\
\paragraph{Modified Architecture for \model}
As shown in Figure~\ref{fig:model_arch}.B, \model~employs a Decoupled Prior Variational Autoencoder (dpVAE) style architecture \cite{bhalodia2020dpvaes}, which combines a VAE with normalizing flows in the latent space.
The DGCNN encoder \(\phi\) maps each input surface \(\X_n\) to its latent representation \(\z\). Specifically, the encoder predicts the mean and standard deviation-\(\mu_{\z},\sigma_{\z}\) of the variational posterior distribution \(q_{\phi}(\z|\X)\) that approximates the true posterior \(p(\z|\X)\). The latent variable \(\z\) is then sampled using the reparameterization trick, defined as: \(\z = \mu_{\z} + \epsilon \odot \sigma_{\z}\) where \(\epsilon\) is a random variable drawn from a standard normal distribution, and \(\odot\) denotes element-wise multiplication. This approach ensures differentiability, enabling efficient backpropagation during training. The NF (\(g_\eta\)) transforms \(\z\) into the sampling space \(\z_0\), facilitating a structured prior for probabilistic modeling. An IM-Net decoder (\(\theta\)) predicts correspondences \(\C\) by using \(\z\) to deform the template to accurately represent the input shape \(\X\).
\\

\paragraph{Training Objective of \model} maximizes the likelihood of the observed data using the dpVAE framework:  
\begin{equation}\label{dpvae_objective}
\begin{aligned}
\mathcal{L}(\theta, \phi, \eta) &= - \mathbb{E}_{\z \sim q_{\phi}(\z|\X)} \left[ \log p_{\theta}(\C|\z) \right] + \\
& \text{KL}\left(q_{\phi}(\z |\X) \| p_{\eta}(\z)\right),
\end{aligned}
\end{equation}
where \(q_{\phi}(\z|\X)\) is the approximate posterior from the encoder, and \(p_{\eta}(\z_0)\) is the prior distribution transformed by the normalizing flow. The likelihood term quantifies reconstruction accuracy using the Chamfer distance between the predicted correspondences \(\C_n\) and the mesh vertices \(\V_n\):  
\begin{equation} \label{chamfer_likelihood}
    -\mathbb{E}_{\z \sim q_{\phi}(\z|\X)}\left[\log p_{\theta}(\C|\z)\right] = \sum_{n=1}^N \mathcal{L}_{\text{Chamfer}}(\V_n, \C_n)
\end{equation}

\noindent\paragraph{Advantages of the Approach} 
\begin{itemize}
    \item Introducing a dpVAE-based framework of VAE with NF enhances the model's ability to capture high-dimensional shape variations, creating a flexible yet structured latent space. 
    
    \item By replacing the SP-VAE, the framework eliminates the need for complex hyperparameter tuning and mitigates issues like mode collapse, particularly in small datasets. This refinement simplifies the training process, enabling efficient end-to-end optimization while ensuring robustness in capturing population-specific shape characteristics.
    
    \item The probabilistic formulation of \model~provides a low-dimensional latent space \(\z\) that facilitates efficient analysis of population statistics. Statistics can be performed directly on \(\z\), replacing the SP-VAE and streamlining the workflow. For generating new samples, latent representations are drawn from the prior \(p(\z_0)\), transformed via the inverse flow mapping \(g_\eta^{-1}\) to obtain \(\z\), and decoded to generate correspondences \(\C\). This iterative sampling process ensures the generated shapes are consistent with the learned shape distribution, and the mean of the generated samples is used as a robust, data-informed template.

\end{itemize}

\item \textit{Surface Projection:}\label{surface_proj} 
Chamfer distance alone does not ensure that predicted correspondences lie precisely on the mesh surface. It only minimizes the average point-to-point distances between predicted and ground truth point clouds. This optimization can result in points that approximate the overall shape but float off the actual surface, especially in areas with complex geometry. Furthermore, Chamfer distance's focus on closest point matches may lead to inaccurate surface representations, particularly when dealing with unevenly distributed point clouds or intricate surface details. Therefore, we introduce a surface projection step to ensure the anatomical accuracy of predicted correspondences. This step aligns the predicted correspondences \(\C\) precisely onto the surface of the input mesh \(\X\), enabling end-to-end training. Given an input mesh \(\X = \{\V, \set{E}\}\) with vertices \(\V \in \mathbb{R}^{K \times 3}\), and predicted correspondences \(\C \in \mathbb{R}^{M \times 3}\), we define the projection process as follows:

1. Compute pairwise distances: Calculate the Euclidean distance between each correspondence point \(\ci \in \C\) and each mesh vertex \(\ve \in \V\):
\begin{equation}
    D_{ij} = \|\ci_i - \ve_j\|_2
\end{equation}
where \(D \in \mathbb{R}^{M \times K}\) is the resulting distance matrix.

2. Calculate softmin weights: To facilitate smooth projection, compute softmin weights for each correspondence point with respect to the vertices:
   \begin{equation}
   W_{ij} = \frac{\exp(-D_{ij} / \sigma)}{\sum_{k=1}^K \exp(-D_{ik} / \sigma)}
   \end{equation}
   where \(\sigma > 0\) controls the softness of the projection.
3. Compute weighted displacements: Using the softmin weights, calculate the displacement vector for each correspondence point:
   \begin{equation}
   \Delta_i = \sum_{j=1}^K W_{ij} (\ve_j - \ci_i)
   \end{equation}
4. Update correspondences: Compute the projected correspondence locations by adding the displacement vectors to the initial correspondences:
   \begin{equation}
       \ci_i^{\text{proj}} = \ci_i + \Delta_i
   \end{equation}
The updated correspondences \(\ci_i^{\text{proj}}\) are then used in the Chamfer distance calculation (Eq. \ref{chamfer_likelihood}).

\item \textit{Vertex Masking and Perturbation:} \model~employs vertex masking and perturbations as a self-supervised learning approach for effective data augmentation. By randomly masking vertices and introducing small perturbations, we challenge the model to reconstruct complete, accurate shapes from partial or noisy inputs. This process generates diverse training examples and encourages the model to learn rich, meaningful shape representations. 

\item \textit{Aleatoric Uncertainty Estimation:} \model~leverages the probabilistic formulation of the modified M-AE with NF to estimate aleatoric uncertainty in predicted correspondences. Unlike epistemic uncertainty (which stems from model limitations), aleatoric uncertainty arises from inherent data noise or ambiguity. This estimation is crucial for assessing the reliability of predictions in different regions of the shape. The process quantifies aleatoric uncertainty as the variance of the conditional distribution \(p(\C_n|\z_n)\). Mathematically, we:
\begin{enumerate}
    \item For a given input mesh \(\X_n\), we sample multiple latent encoding: \(\z^{(s)}_n\sim \mathcal{N}(\z_n|\mu_{\z},\sigma_{\z})\), and \(s=1,\ldots,S\) represents the samples
    \item Get the correspondences for all the samples: \(\C^{(s)}_n=f_{\theta}(\z^{(s)}_n)\)
    \item Fit a Gaussian distribution to the decoded predictions: \(\mathcal{N}(\C_n|\mu,\sigma)\)
\end{enumerate}
The variance \(\sigma^2\) of this fitted Gaussian represents the aleatoric uncertainty, highlighting regions of higher prediction ambiguity or noise. This approach assumes that the prediction distribution is approximately Gaussian, which may not always hold. However, it provides a computationally efficient way to estimate uncertainty, enabling more informed decision-making in downstream tasks such as shape analysis or reconstruction.

\end{enumerate}
These enhancements collectively overcome the challenges faced in Mesh2SSM, including the difficulties in training SP-VAE and the complexities of the loss function. By integrating a bidirectional flow, simplifying the loss calculation, and adding a surface projection step, \model~provides a more robust, efficient, and anatomically accurate solution for statistical shape modeling from meshes.

\section{Experiments}
This section presents a comprehensive evaluation of our proposed \model~method. We begin by detailing the evaluation metrics used to assess the quality of the shape models, covering surface sampling accuracy, correspondence quality, and SSM performance. Following this, we describe the diverse anatomical datasets employed in our experiments, highlighting their unique characteristics and challenges. We then introduce the comparison models, including SOTA methods in shape modeling and analysis. 

\subsection{Evaluation Metrics} \label{metrics}
This section outlines the metrics used to evaluate the quality of the shape models. 
\begin{enumerate}
    \item \textit{Surface sampling metrics to assess accuracy of surface representation}
    \setlist[itemize]{leftmargin=0pt}
    \begin{itemize}
        \item \textbf{Chamfer Distance (CD)} measures the average distance between two point sets, calculated bidirectionally: from each point in set \(\C_j\) to its nearest neighbor in set \(\V_j\), and vice versa. This provides a comprehensive measure of dissimilarity between the two point sets.
        \item \textbf{Point-to-Mesh Distance (P2M)} is determined by summing two components: the point-to-mesh face distance and the face-to-point distance. This distance is calculated between the predicted correspondences \(\C_j\) and the mesh defined by vertices \(\V_j\) and edges \(\set{E}_j\).
    \end{itemize}
    \item \textit{Correspondence metrics to assess ability to capture population-level statistics:}
    \begin{itemize}
        \item \textbf{Surface-to-Surface Distance (S2S)} is computed between the original surface mesh and a generated mesh derived from predicted correspondences. To obtain the reconstructed mesh, correspondences are matched to the mean shape, and the warp between the predicted correspondences and the mean particles is applied to the mean mesh to get the reconstructed surface. Smooth reconstruction and low surface-to-surface distance indicate good quality of correspondences.
    \end{itemize}
    \item \textit{SSM Metrics:} evaluate the quality and performance of the shape models, ensuring they accurately represent the shape variations in a population while maintaining the ability to describe new instances and generate valid shapes.
    \setlist[itemize]{leftmargin=0pt}
    \begin{itemize}
    \item \textbf{Compactness} measures how efficiently a shape model represents the variability in a population using as few parameters as possible. It is quantified by the number of principal components (PC) or shape modes needed to explain a certain percentage of the total shape variance. A more compact model requires fewer PCs to capture the same variation. Mathematically, compactness can be defined as the cumulative explained variance of the Mth eigenmode obtained by the model's covariance matrix decomposition.

    \item \textbf{Generalization} assesses how well the shape model can describe shapes that were not part of the training set. It evaluates the model's ability to represent new, unseen instances of the shape class. This is typically measured by the reconstruction error when the model attempts to match new data. A lower reconstruction error indicates better generalization.

    \item \textbf{Specificity} measures the model's ability to generate valid instances of the trained shape class. It quantifies how well the shapes generated by the model resemble those in the training set. This is often calculated as the average distance between randomly sampled model-generated shapes and the nearest shapes in the training set. A lower average distance indicates better specificity. 
    \end{itemize}
\end{enumerate}
\subsection{Dataset}
We employ segmentation datasets (three public and two in-house) comprising five distinct anatomies with varying cohort sizes. All datasets undergo random partitioning into training, validation, and testing sets using an \(80\%/10\%/10\%\) split. Each dataset is described as follows:
\begin{itemize}
\item \textbf{Femur (56 shapes)}: The femur dataset contains proximal femur bones clipped under the lesser trochanter to focus on the femoral head. Nine of the femurs have the cam-FAI pathology characterized by an abnormal bone growth lesion that causes hip osteoarthritis. 
\item \textbf{Spleen (40 shapes)} \cite{simpson2019large}: The spleen organ dataset provides a limited data scenario with challenging shapes that vary significantly in size and curvature.
\item \textbf{Pancreas (272 shapes)} \cite{simpson2019large}: The pancreas dataset comprises of pancreas organs with tumors of varying sizes from cancer patients, providing complex patient-specific shape variability.
\item \textbf{Liver (834 shapes)} \cite{Ma-2021-AbdomenCT-1K}: The liver dataset provides an organ dataset with nonlinear shape variation.
\item \textbf{Left Atrium (923)}: This dataset includes 3D late gadolinium enhancement (LGE) and stacked cine cardiovascular magnetic resonance (CMR). The dataset comprises 923 anonymized obtained from distinct patients and was manually segmented by cardiovascular medicine experts at the University of Utah Division of Cardiovascular Medicine; the endocardium wall was used to cut off pulmonary veins.
\end{itemize}

\subsection{Comparison Models}
\begin{itemize}
\item \textbf{ShapeWorks (SW)} \cite{cates2017shapeworks} is a SOTA PSM tool that establishes dense correspondences on complete surface representations using a particle-based approach. It is computationally efficient and widely adopted for shape variability analysis. We use its open-source implementation to benchmark our method against standard SSM metrics.
\item \textbf{Deformetrica} \cite{durrleman2014morphometry}, a large deformation diffeomorphic metric mapping (LDDMM) framework, is a well-established method in the medical imaging domain and serves as a baseline for state-of-the-art techniques. Unlike data-driven approaches, LDDMM formulates shape alignment as a pairwise optimization problem. Its objective is to minimize the varifold distance between the target shape surface mesh and the template surface mesh. We use the open-source implementation of Deformetrica, and only the kernel width parameter was altered to gain better results on each dataset; otherwise, we use the default parameters.
\item \textbf{FlowSSM} \cite{ludke2022landmark} operates directly on surface meshes and uses neural networks to parameterize the deformations field between two shapes in a low dimensional latent space and rely on an encoder-free setup. The encoder-free step randomly initializes the latent representations for each sample, and the latent representations are optimized to produce the optimal deformations.
\item \textbf{Mesh2SSM} \cite{iyer2023mesh2ssm} model described in section~\ref{mesh2ssm}
\item \textbf{\model~*} autoencoder version without normalizing flows and simple template update as the average of training correspondences. Abbreviated as M++AE for readability. 
\item \textbf{\model} is the proposed model described in section~\ref{mesh2ssm_plus} and abbreviated as M++Flow for readability. 
\end{itemize}
All models use the same median-shape template mesh to avoid bias. \model-based approaches periodically update the template based on learned statistics. In the case of the M++Flow model, the template is updated using the sampling procedure described in section~\ref{mesh2ssm_plus}. In contrast, for M++AE, the template is updated as the mean correspondence of all the predicted training samples. All SW, Mesh2SSM, and \model~based methods use 1024 correspondence points, whereas Deformetrica and FlowSSM establish vertex-wise correspondences. Note that we attempted to replicate the results of the FUSS \cite{el2024universal} using the code and hyperparameters provided by the authors. However, we were unable to achieve the performance reported in their paper \cite{el2024universal}. To ensure fairness, we have excluded the FUSS model from our comparative analysis.

\section{Results}
\subsection{Quantitative and Qualitative Analysis}
\begin{figure}
    \centering
    \includegraphics[width=\linewidth]{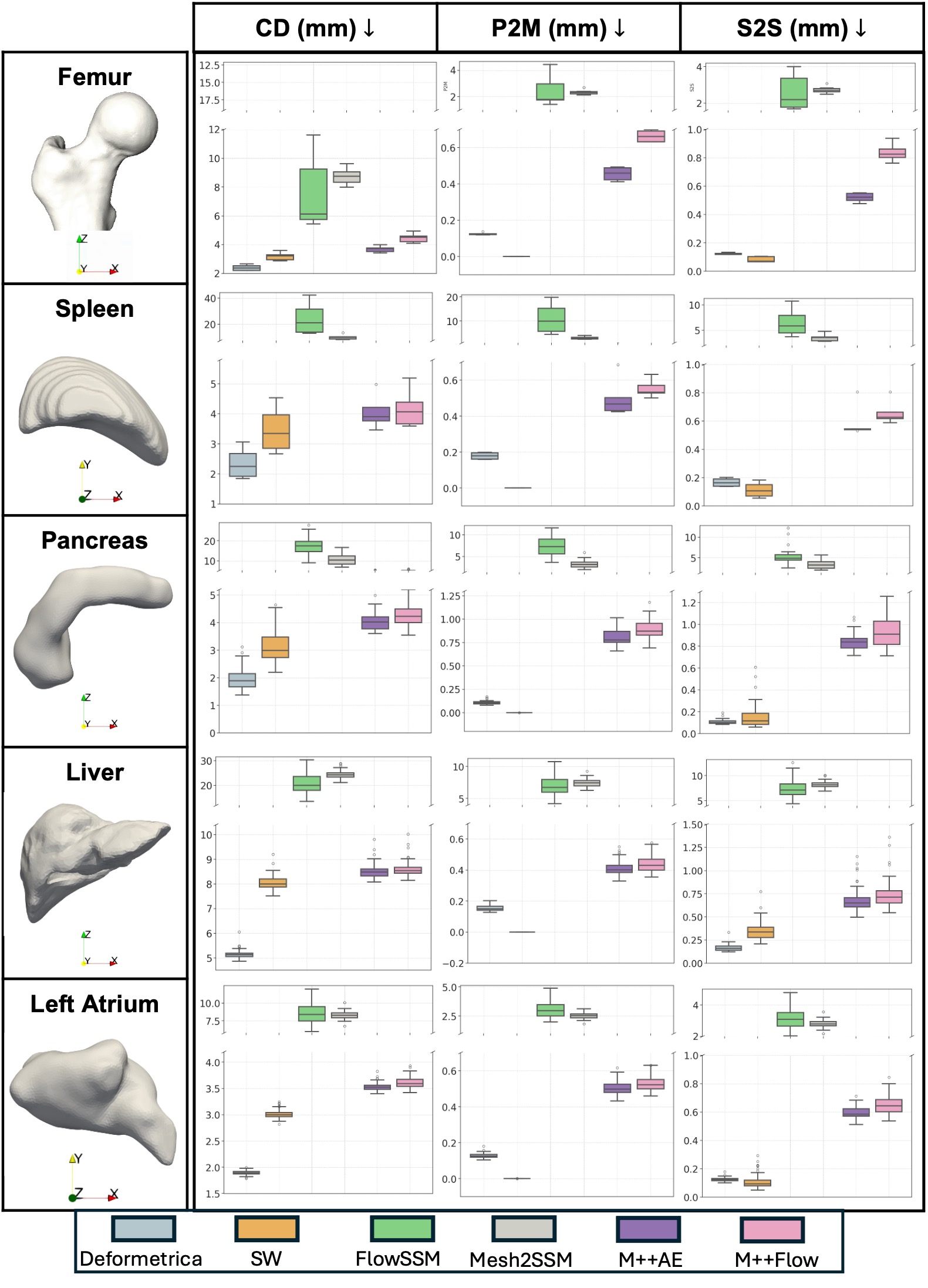}
    \caption{\textbf{Distance Metrics:} Boxplots show the error distribution across test sets for each model in mm. } 
    \label{fig:distance_metrics}
\end{figure}

Figure~\ref{fig:distance_metrics} compares five methods—Deformetrica, SW, FlowSSM, Mesh2SSM, and the \model~based models (M++AE and M++Flow)—across five anatomical structures using CD, P2M, and S2S distance metrics. SSM and Deformetrica are included as reference points for these metrics. Lower values across all metrics indicate better performance.

\paragraph{CD:} The femur and spleen, characterized by their simple anatomical structures, direct optimization-based methods Deformetrica and SW achieve the lowest CD values, while the \model~based models closely follow, demonstrating high reconstruction accuracy. For more complex anatomies, such as the liver and left atrium, the \model~based models perform comparably to traditional optimization methods, underscoring their ability to model intricate geometries effectively. In contrast, FlowSSM and Mesh2SSM display higher CD values, particularly for the pancreas and liver datasets.

\paragraph{P2M} Consistent with the CD metric, Deformetrica, SW, and the \model~based models consistently achieve accuracies close to 1 mm, outperforming Mesh2SSM and FlowSSM. These results highlight the adaptability of the \model~based models in faithfully representing complex shape surfaces.

\paragraph{S2S} M++Flow and M++AE exhibit lower S2S values than other deep learning-based methods for all datasets and achieve great performance with \(<=1 mm\) error, showcasing their robustness in handling intricate surface details.

These findings underscore the importance of selecting appropriate shape modeling methods based on the complexity of the target anatomy. The proposed methods, M++Flow and M++AE, demonstrate clear advantages, mainly as they learn to represent shapes using a minimal number of correspondences—often fewer than the number of vertices. In contrast, Deformetrica and FlowSSM require vertex-wise correspondences, complicating the training and inference process. Additionally, the ability of the \model~based models to update the template, perform surface projections, and incorporate data augmentation steps during training helps mitigate overfitting, a common issue observed in other deep learning approaches.
\begin{figure}
    \centering
    \includegraphics[width=\linewidth]{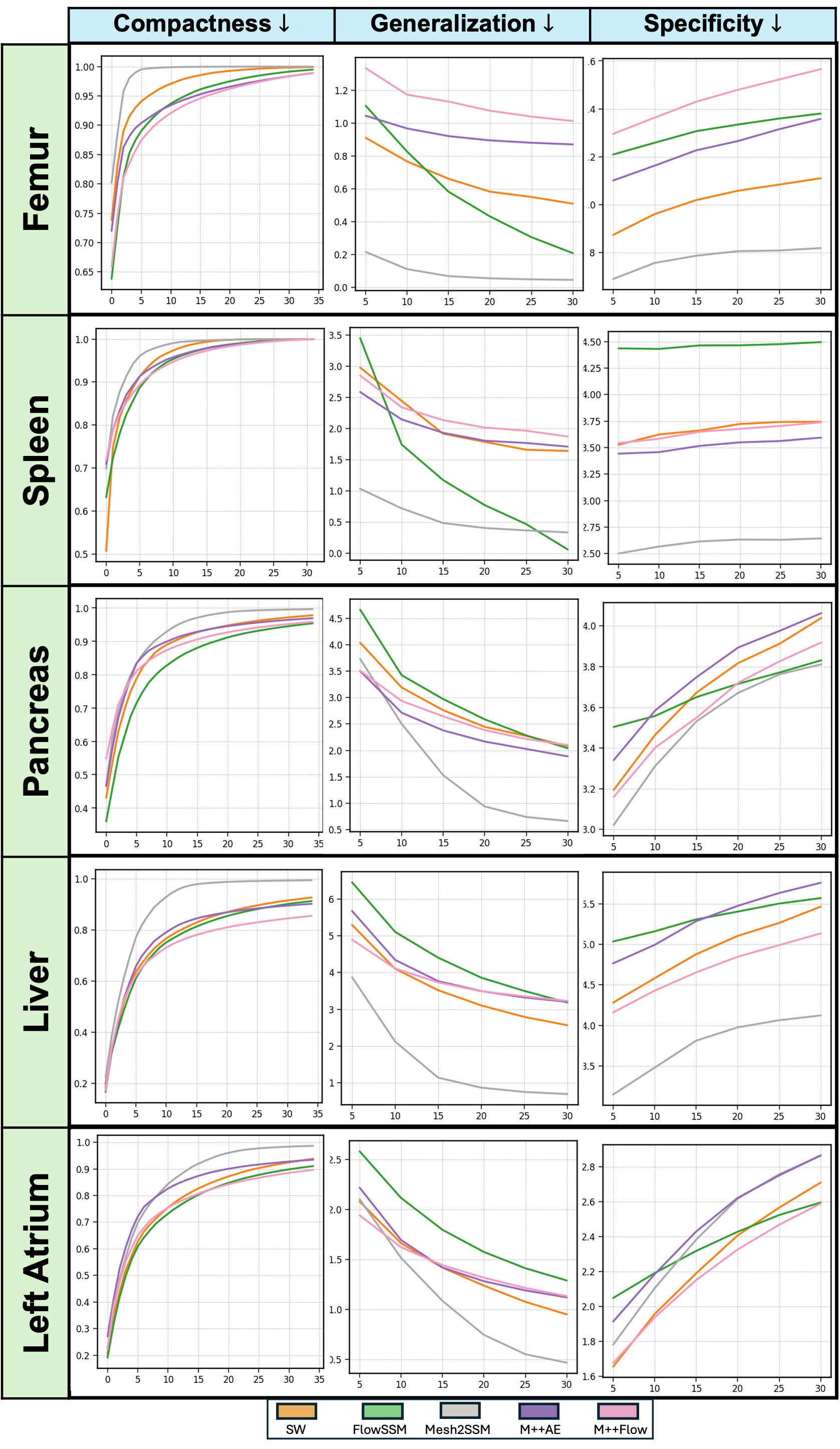}
    \caption{\textbf{SSM Metrics: }A compactness plot displays the cumulative variance ratio as a function of PCA mode count. Generalization and specificity reconstruction error plotted as a function of PCA mode count. For all datasets, a maximum of 30 modes that account for at least 99\% of the total variation are shown.}
    \label{fig:ssm_metrics}
\end{figure}

\paragraph{SSM Metrics} Figure~\ref{fig:ssm_metrics} illustrates the SSM metrics (described in section \ref{metrics}) plotted as a function of PCA mode count. Deformetrica was excluded from the SSM metrics plots due to its high computational demands, particularly for large datasets like the left atrium and liver. Its reliance on LDDMM \cite{durrleman2009statistical} ensures high-quality shape representations, but this comes at the cost of significant computational resources \cite{bautz2023unsupervised}, making it less practical for large-scale comparisons. Deformetrica was not included in these plots to maintain uniformity and focus on scalable methods. The figure demonstrates that the performance of the \model~based models matches or exceeds that of the traditional PSM optimization method ShapeWorks on SSM metrics.
In most cases, the \model~based models achieve similar or better compactness and better generalization and specificity. While Mesh2SSM and FlowSSM may exhibit lower generalization and specificity and higher compactness in some instances, SSM metrics alone do not provide a complete assessment of model quality. It is essential to consider the distance metrics from Figure~\ref{fig:distance_metrics} alongside the SSM metrics in Figure~\ref{fig:ssm_metrics}.

\begin{figure}
    \centering
    \includegraphics[width=\linewidth]{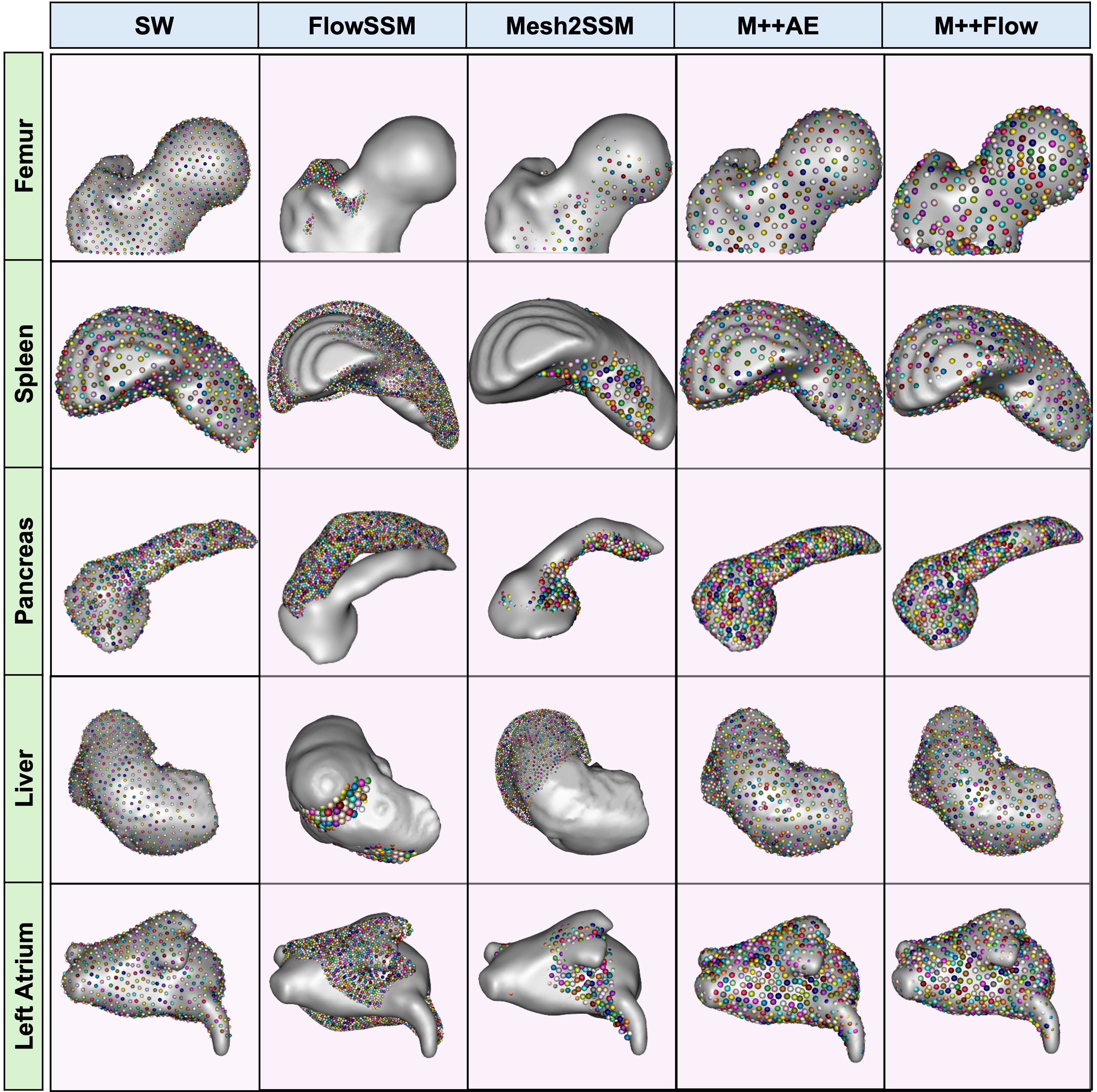}
    \caption{\textbf{Correspondence Quality: }Predicted correspondence
point for test meshes are overlaid over ground truth meshes for all methods and datasets.}
    \label{fig:corr_quality}
\end{figure}

This observation is further supported by the qualitative results in Figure~\ref{fig:corr_quality}, which display the predicted correspondences for test samples across all methods and datasets. Models prone to overfitting often yield good reconstruction for training samples, resulting in favorable SSM metrics, but their distance-based metrics on test samples reveal poorer performance. Specifically, overfitting in the case of FlowSSM may be influenced by the complexity and sensitivity of its hyperparameters and the encoder-free setup, which requires the inference-time optimization of the latent encoding, making them more challenging to tune for diverse datasets. SW and \model~based methods produce high-quality correspondences uniformly distributed across the surface of the ground truth meshes. Although Mesh2SSM also generates well-distributed correspondences, these do not always align with the surface of the mesh. This limitation has been addressed in the proposed \model~through the incorporation of surface projection, as detailed in Section~\ref{mesh2ssm_plus}.

\begin{table*}
\centering
\caption{\textbf{Statistical Correlations: } Pearson and Spearman correlation with p-values between aleatoric uncertainty and CD/P2M for all datasets. Statistically significant correlations (\(p < 0.05\)) are highlighted in bold.}
\label{tab:sample_wise_correlation}
\vspace{-2mm}
\footnotesize 
\begin{tabular}{|c|c|c|c|c|c|c|}
\hline
\textbf{Quantity} & \textbf{Metric} & \textbf{Spleen} & \textbf{Femur} & \textbf{Liver} & \textbf{Pancreas} & \textbf{Left Atrium} \\ \hline
\multirow{2}{*}{CD} 
& Pearson & \textbf{0.9692 (0.0064)} & 0.6816 (0.1359) & \textbf{0.4260 (4.8e-5)} & \textbf{0.6379 (1.97e-4)} & \textbf{0.3859 (1.33e-4)} \\ 
& Spearman & \textbf{0.9000 (0.0374)} & 0.7143 (0.1108) & \textbf{0.2817 (0.0090)} & \textbf{0.7187 (1.13e-5)} & \textbf{0.3708 (2.53e-4)} \\ \hline
\multirow{2}{*}{P2M} 
& Pearson & -0.8700 (0.0552) & -0.4223 (0.4042) & \textbf{0.3935 (1.95e-4)} & \textbf{0.3767 (0.0440)} & \textbf{0.5687 (2.73e-9)} \\ 
& Spearman &\textbf{ -0.9000 (0.0374)} & -0.4286 (0.3965) & \textbf{0.3071 (0.0043)} & 0.2463 (0.1977) & \textbf{0.5184 (1.02e-7)} \\ \hline
\end{tabular}

\end{table*}

\subsection{Aleatoric Uncertainty and Error Correlations}
\begin{figure}
    \centering
    \includegraphics[width=\linewidth]{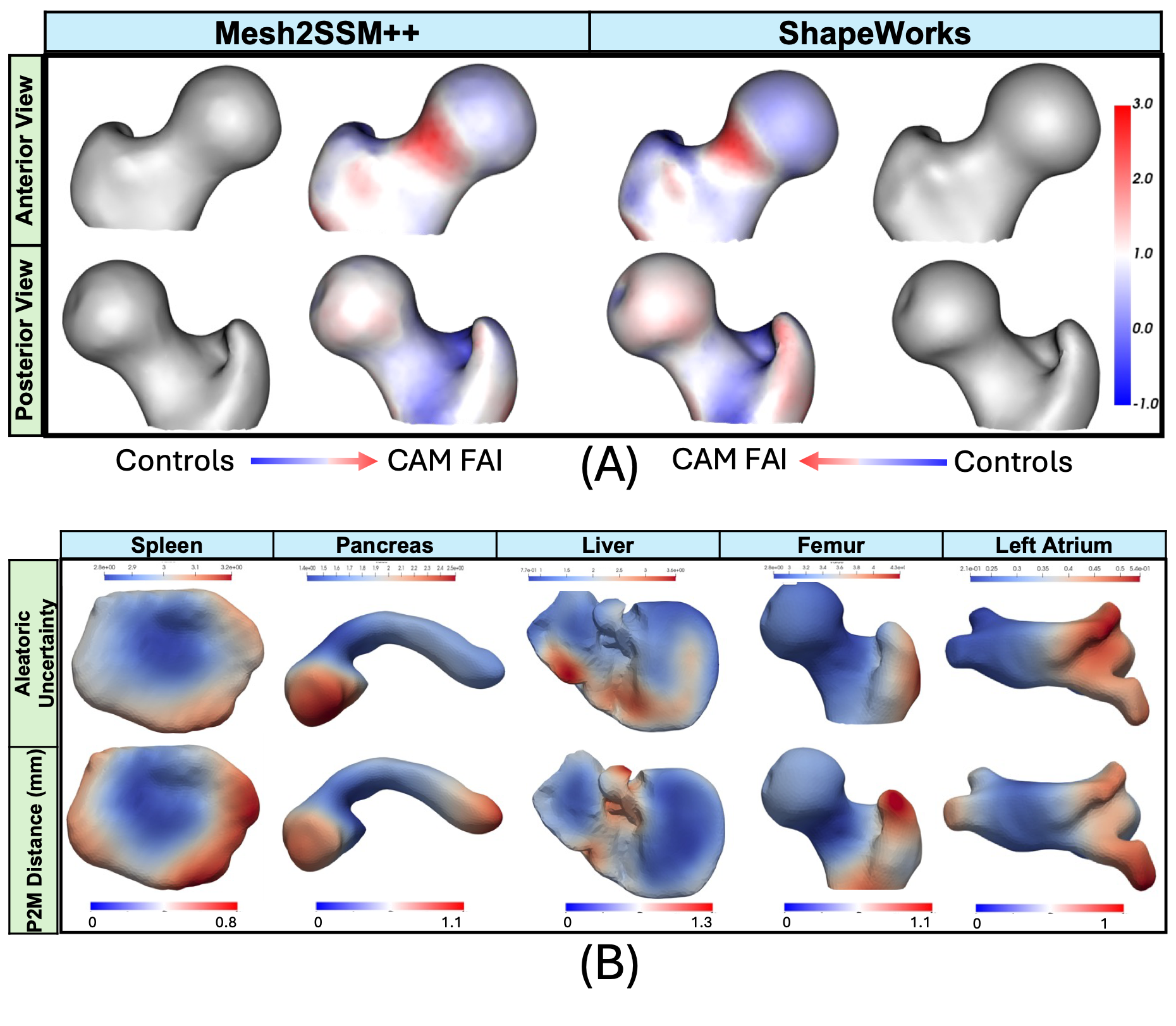}
    \caption{\textbf{(A) Comparison of Group Differences Identified by ShapeWorks vs. \model:} The figure illustrates the mean shapes of the control group. Color mapping indicates the distance between the control and CAM FAI mean. Both ShapeWorks and \model successfully capture the characteristic widening of the femoral neck associated with the CAM FAI pathology. \textbf{(B) Uncertainty Calibration:} Heatmaps on a representative mesh display average P2M error and aleatoric uncertainty, highlighting spatial correlation.}
    \label{fig:group_diff_uncertainity_surface}
\end{figure}
Figure~\ref{fig:group_diff_uncertainity_surface}.B presents the correlation between aleatoric uncertainty and CD and P2M distance errors across five datasets. The spatial correlation between uncertainty and error heatmaps underscores the utility of probabilistic frameworks in assessing prediction reliability. Regions with higher uncertainty values correspond to areas where predicted points exhibit greater deviation from the true surface. Table~\ref{tab:sample_wise_correlation} quantifies the sample-wise and particle-wise correlations using Pearson and Spearman coefficients.

The spleen dataset demonstrates the strongest correlation, particularly for CD (Pearson: \(0.9692, p = 0.0064\)), highlighting that uncertainty effectively captures surface deviations, as visualized in Figure~\ref{fig:group_diff_uncertainity_surface}.B. Although the spleen is a relatively simple shape, the cohort exhibits high variability, and the small dataset size further emphasizes the importance of accurate uncertainty quantification. 
Conversely, the femur dataset shows weak and statistically insignificant correlations (\(p > 0.05\)), likely due to its simpler geometry and smaller surface deviations and uncertainty variations. The liver dataset reveals significant correlations for CD and P2M, suggesting that uncertainty estimation is well-calibrated. Similarly, the pancreas and left atrium datasets exhibit significant positive correlations, with higher uncertainty corresponding to larger errors, particularly for P2M distances, as corroborated by Figure~\ref{fig:group_diff_uncertainity_surface}.B.

These results indicate that aleatoric uncertainty is well-calibrated in most datasets, particularly regions with more significant surface deviations. The spatial correlation between uncertainty and error maps underscores the critical role of probabilistic frameworks in evaluating model reliability. This is especially valuable for complex or irregular shapes, highlighting the potential of such frameworks to enhance robustness in clinical applications where reliable uncertainty estimation supports informed decision-making.

\subsection{Outlier Detection}
\begin{figure}
    \centering
    \includegraphics[width=\linewidth]{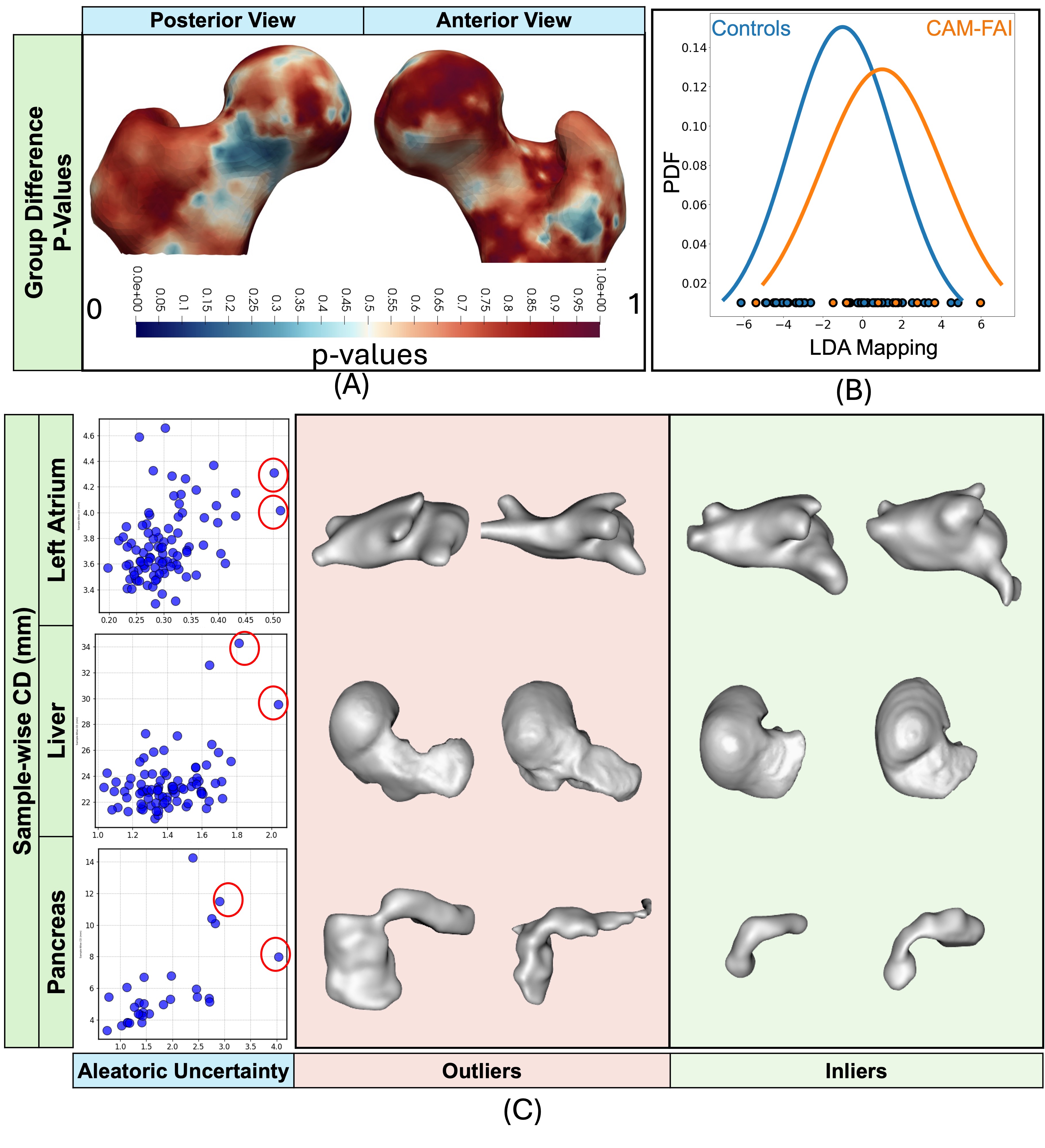}
    \caption{\textbf{A. Group Difference Statistical Significance:} The p-values of the group differences overlayed over the mean mesh. The color showcases statistical significance. \textbf{(B). LDA Map} Shape mapping to linear discrimination of variation between population means for the groups of patients and controls. \textbf{(C) Aleatoric Uncertainty for Outlier Detection: }Calibrated aleatoric uncertainty scatter plots against sample-wise CD. Detected outliers with the highest aleatoric uncertainties and CD are indicated in red. The two inlier samples with the lowest aleatoric uncertainties and the identified outliers are compared. }  
    \label{fig:lda_outliers}
\end{figure}

Figure~\ref{fig:lda_outliers}.C presents scatter plots of aleatoric uncertainty against CD, illustrating the utility of uncertainty quantification in identifying out-of-distribution (OOD) samples. For instance, the scatter plot highlights two outliers from each dataset with a high CD and aleatoric uncertainty, marked in red and visualized in Figure~\ref{fig:lda_outliers}.C. These outliers exhibit irregular shapes and significant variability compared to inliers, with lower uncertainties and errors.
In the pancreas dataset, the original shapes are derived from manually segmented CT scans, as described in \cite{simpson2019large}. The ambiguity in labeling this small organ often results in poor-quality surface meshes. Aleatoric uncertainty calibration effectively identifies such outliers, characterized by unusually thin structures and high variability. Similarly, in the left atrium dataset, aleatoric uncertainty highlights shapes with thin, less ellipsoid-like structures compared to the more uniform morphology of inliers. These variations may indicate structural remodeling or atrophy associated with different diseases \cite{casaclang2008structural}. In the liver dataset, the identified outliers display an elongated, thinning left lobe, which could suggest chronic liver damage, such as cirrhosis or fibrosis \cite{higaki2023liver}. The ability of aleatoric uncertainty to flag such morphologically distinct outliers demonstrates its value in detecting anomalies that may have clinical significance.

\subsection{Modes of Variations}
\begin{figure}
    \centering
    \includegraphics[width=\linewidth]{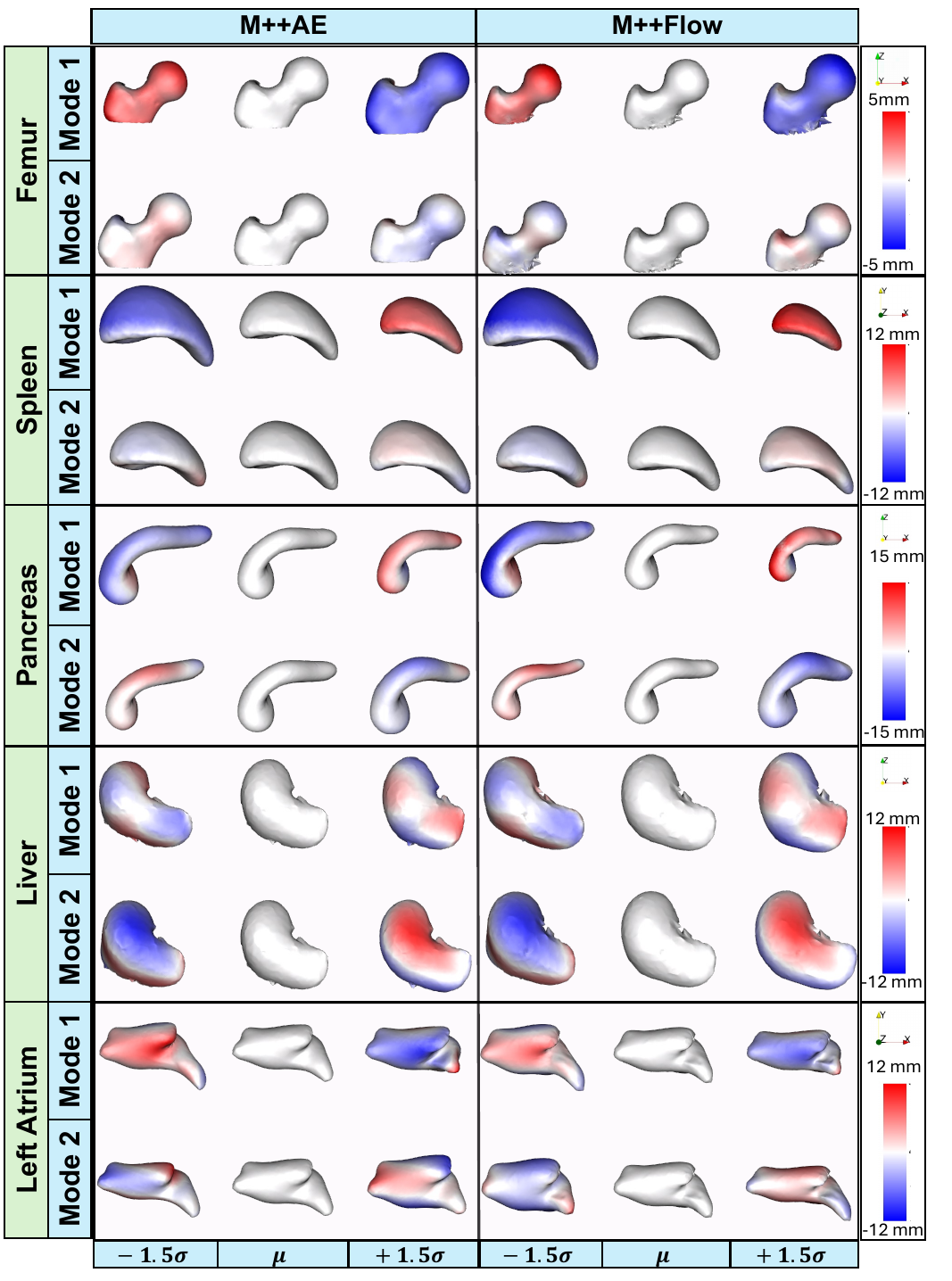}
    \caption{\textbf{Modes of Variation:} The first two modes were identified by performing PCA on the predicted correspondences. } 
    \label{fig:data_pca}
\end{figure}
\begin{figure}
    \centering
    \includegraphics[width=\linewidth]{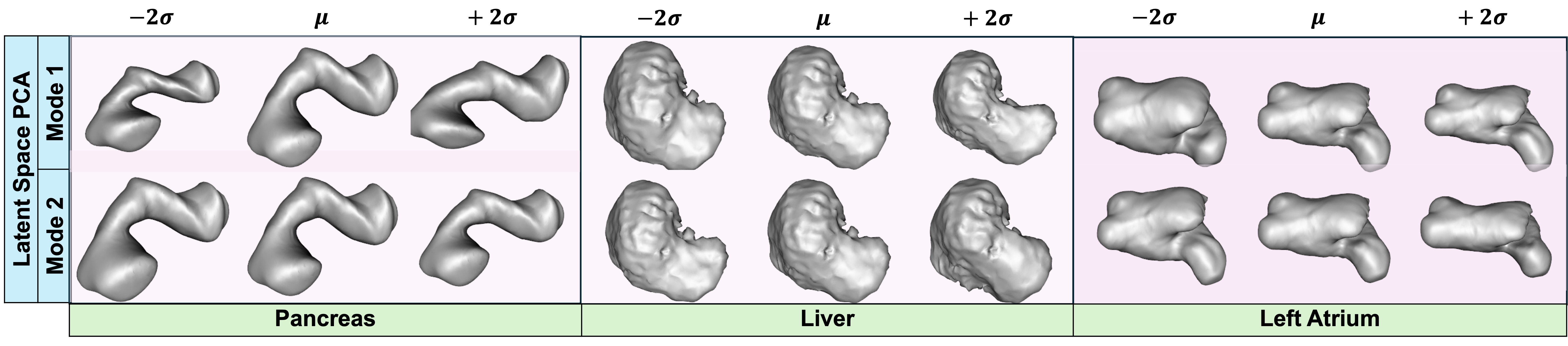}
    \caption{\textbf{Modes of Variation in Latent Space:} The first two modes were identified by performing PCA in the latent space of the M++Flow model for the liver, left atrium, and pancreas datasets.}
    \label{fig:latent_pca}
\end{figure}

Figure~\ref{fig:data_pca} depicts the top two PCA modes of variation for all datasets identified by the \model~based models. Both models identify similar modes of variation, suggesting that the addition of normalizing flows does not compromise the quality of the correspondences. By incorporating a normalizing flow in the latent space of the M++Flow model, we can generate new samples and utilize an additional representation space—beyond the correspondences—for statistical analysis. To explore this further, we performed PCA on the learned latent representations of the pancreas, liver, and left atrium datasets. The top two modes of variation in the latent space are shown in Figure~\ref{fig:latent_pca}.

For the pancreas dataset, the first mode represents the curvature of the anatomy, while the second mode reflects the overall size of the pancreas and the roundness of its head. The pancreas dataset contains samples from pancreatic cancer patients, typically present in the head of the pancreas. The PCA modes in Figure~\ref{fig:data_pca} and Figure~\ref{fig:latent_pca} effectively capture these population-level variations. In the liver dataset, the first mode of variation, derived from the latent space and correspondence-based representations, highlights the thinning of the liver's left lobe—a well-documented population-level variation in liver morphology. Similarly, for the left atrium dataset, the modes of variation identify the roundness of the organ and the length of the pulmonary veins as key contributors to anatomical diversity.

\subsection{Downstream Task Analysis}
\paragraph{Group Differences}
The femur dataset consists of nine subjects with CAM pathology, a condition characterized by aberrant bone development on the femoral neck that restricts motion and is associated with hip osteoarthritis, and 47 healthy/control subjects. Figure~\ref{fig:group_diff_uncertainity_surface}.A presents the group differences analysis using the predicted correspondences. The \model~based model captures the distinction between CAM pathology and control subjects and is comparable to the conventional PSM technique SW. The color map represents the distance between the mean shapes of the controls and the pathology group. This experiment demonstrates how \model~can be used for pathology localization.
\begin{figure}
    \centering
    \includegraphics[width=\linewidth]{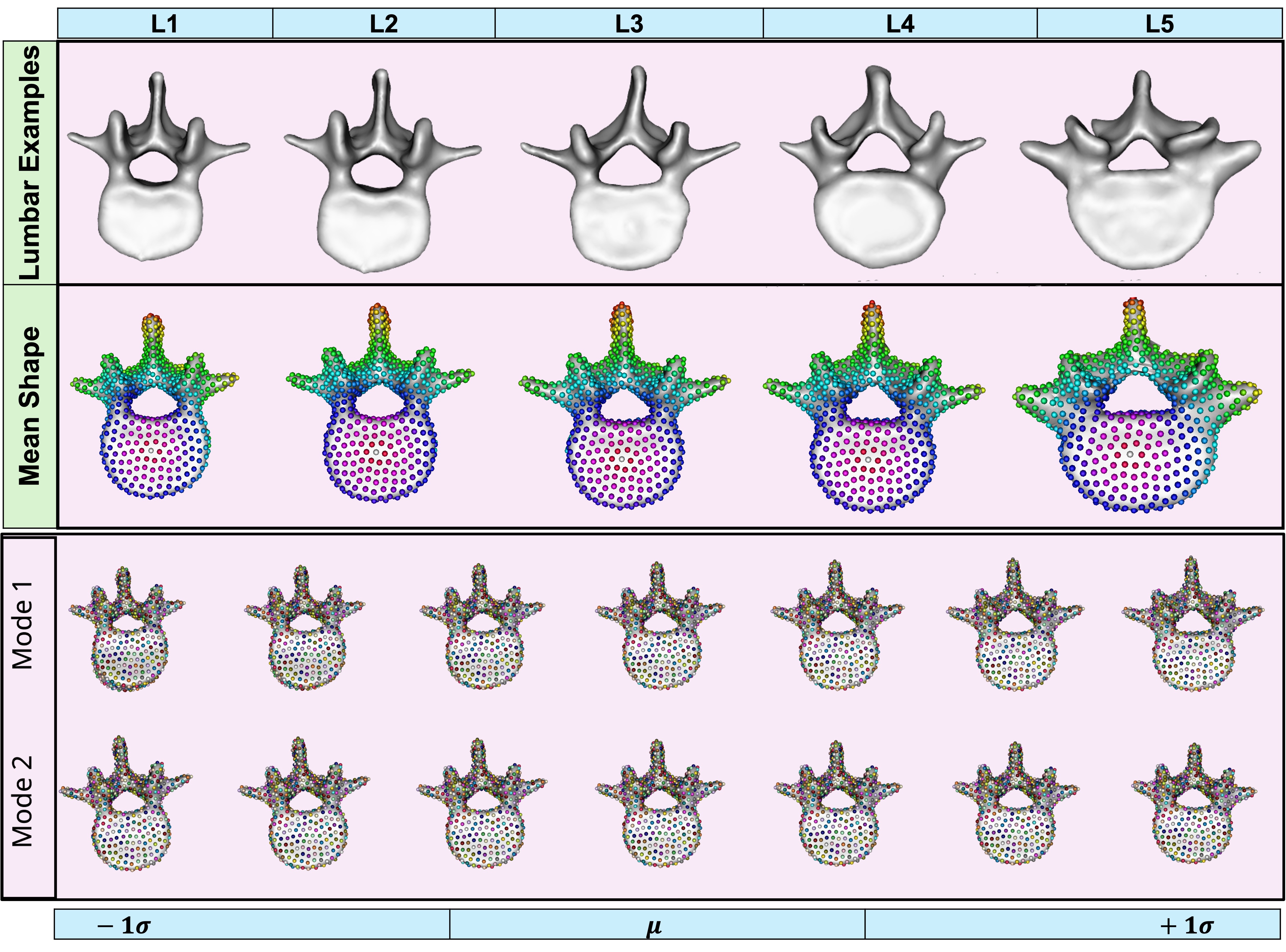}
    \caption{\textbf{Lumbar Vertebra: } The first row represents the examples of the lumbar vertebra. The second row represents the mean predictions from the \model~model for each vertebrae. Color denotes correspondence. The next two rows represent the PCA modes of variations identified by \model that encompasses all five categories of the lumbar vertebra.}
    \label{fig:lumbar}
\end{figure}
Additionally, we evaluated the statistical significance of the estimated group differences using the Hotelling metric with a nonparametric permutation test and false discovery rate (FDR) correction \cite{cates2007shape} for multiple comparisons. This approach enables the identification and visualization of localized regions with significant shape differences. The null hypothesis asserts that the distributions of corresponding sample points are identical across groups. Lower p-values \((<0.05)\) indicate rejection of the null hypothesis, suggesting that the observed group differences are significant and not derived from the same distribution. Figure~\ref{fig:lda_outliers}.A overlays p-values on the mean shape, with blue regions at the head-neck junction of the femur indicating statistically significant differences. These findings align with the regional group difference analysis shown in Figure~\ref{fig:group_diff_uncertainity_surface}.A, further validating the utility of \model for detecting and quantifying meaningful group-level morphological variations.

\paragraph{Linear Discriminant Analysis (LDA)} We employed Linear Discriminant Analysis (LDA) to investigate shape variations between patients with and without CAM impingement and to analyze the distribution of individual shapes within these groups. The mean shape of the CAM impingement group (calculated as the average particle correspondence locations) was compared to the mean shape of the non-CAM group. The linear discriminant vector was defined as the difference between the mean shape vectors. Each subject's shape was mapped onto this discriminant vector by computing the dot product between the subject-specific shape representation (particle correspondences) and the difference vector. This projection produced a scalar "shape-based score" that positioned each subject's anatomy along the group-derived shape difference. To improve interpretability, the mappings of the mean shapes were normalized to -1 (for patients with CAM impingement) and 1 (for controls without CAM impingement). Individual subject mappings were scaled relative to these values, providing a distribution of shape scores across the population, with members clustering near the mean shapes of their respective groups. A univariate Gaussian distribution was fitted to the normalized mappings for each group to define the probability density function of shape scores. Figure~\ref{fig:lda_outliers}.B displays the LDA mappings for controls and CAM pathology samples, showing two distinct distributions. The overlap in these distributions can be attributed to the limited sample size, suggesting that this analysis could be further refined with a larger dataset.

\paragraph{Multi-Anatomy Modeling and Downstream Classification}
A shape model was developed using publicly available, labeled, and segmented data from the vertebral segmentation challenge (VerSe) \cite{sekuboyina2021verse}. Specifically, lumbar vertebrae data were used, including L1 (118 samples), L2 (60 samples), L3 (128 samples), L4 (40 samples), and L5 (119 samples). These samples were split into training, testing, and validation subsets. The \model~was trained using all lumbar vertebrae, with the medoid of the dataset serving as the template. The multi-anatomy setup did not update the template to maintain global anatomical consistency and preserve structural relationships. As the most central shape, the medoid template provided a suitable reference without further refinement. Figure~\ref{fig:lumbar} illustrates the mean shapes produced by \model~for each subgroup. These mean shapes correctly characterize the anatomical distinctions among the vertebrae: L1, the smallest lumbar vertebra, has a compact structure; L2 exhibits a slightly larger body with stronger processes; L3, situated at the midpoint of the lumbar region, has a broader and thicker body for enhanced support; L4 is larger still; and L5, the largest vertebra, features a wedge-shaped body thicker anteriorly.

To explore the utility of SSMs for shape classification, we trained a classifier using the SSM predictions from each construction technique. Correspondences for training and testing samples were first obtained from each model. A multilayer perceptron (MLP) with 100 neurons was trained using five-fold cross-validation, ensuring a fair analysis by repeating the experiment with different train/test splits. The results, summarized in Table~\ref{tab:classification_results}, indicate that all SSMs correctly classified at least 80\% of cases. The low performance of FlowSSM could also be attributed to the model's ability to over-fit, which could lead to sub-optimal correspondences in the test and validation set, thereby reducing the overall classification accuracy of the shape model. On the contrary, \model~based models and SW achieved the highest accuracy at 98\%. These findings highlight that \model~provides SSMs capable of capturing nuanced morphological differences indicative of subtypes, even when shapes are highly similar.

\begin{table}[h!]
\centering
\caption{\textbf{Lumbar Vertebra Classification Performance: } Overall accuracy and F1 score mean and standard deviation along with the class-specific F1 scores using five-fold cross-validation. }
\label{tab:classification_results}
\resizebox{\linewidth}{!}{
\begin{tabular}{|c|c|c|c|c|c|c|c|}
\\\cline{4-8}
 \multicolumn{3}{c|}{} & \multicolumn{5}{|c|}{\textbf{Class F1 Score}} \\ \hline
\textbf{Method} & \textbf{Accuracy} & \textbf{F1-Score} & \textbf{L1} & \textbf{L2} & \textbf{L3} & \textbf{L4} & \textbf{L5} \\ \hline
M++AE       &   0.98 ± 0.02  &  0.97 ± 0.03  &  1.00     & 0.98     & 0.96    & 0.95     & 0.97     \\ \hline
M++Flow     &   0.98 ± 0.02  &  0.97 ± 0.03  &  1.00     & 0.98     & 0.96    & 0.95     & 0.97     \\ \hline
Mesh2SSM    &   0.98 ± 0.01  &  0.98 ± 0.01  &  0.99     & 0.98     & 0.98    & 0.97     & 0.97     \\ \hline 
FlowSSM     &   0.81 ± 0.03  &  0.82 ± 0.03  &  0.73     & 0.85     & 0.86    & 0.83     & 0.82     \\ \hline 
ShapeWorks  &   0.98 ± 0.00  &  0.98 ± 0.00  &  1.00     & 0.98     & 0.98    & 0.95     & 0.93     \\ \hline 
\end{tabular}
}
\end{table}

\section{Limitations and Future Work}
The current model assumes that the cohort of shapes is roughly aligned, which may limit its applicability to diverse datasets and clinical scenarios. Addressing this limitation by developing robust alignment algorithms or exploring alignment-free approaches could significantly expand the usability of \model~across various applications. Furthermore, improving the robustness and computational efficiency of mesh feature extraction methods would eliminate the reliance on geodesic distance calculations, which are often computationally intensive. Instead, leveraging alternative representations that preserve the structural knowledge of meshes—such as topological features, intrinsic coordinates, or spectral embeddings—can balance efficiency and fidelity to the mesh's geometric properties. These approaches enable the model to capture critical shape characteristics without the overhead of geodesic distance computation, paving the way for scalable and robust feature extraction in complex datasets.

While the CD effectively ensures that predicted correspondences are near the ground truth surface, it does not guarantee that they lie exactly on it. Our model addresses this limitation by incorporating a surface projection step, which aligns predicted correspondences directly with the input mesh surface. However, CD may face challenges when the input mesh contains missing regions, spurious surfaces, or significant noise, potentially leading to suboptimal or inaccurate results \cite{araslanovdiffcd}.

Alternative approaches, such as Neural Implicit Functions \cite{wang2023neural}, that leverage self-supervised construction of Signed Distance Fields (SDFs) for surface reconstruction accurately define the underlying surface using the zero-level set. Initially developed for point cloud scenarios, Neural Implicit Functions can be adapted to address mesh irregularities effectively. This extension could improve generalization, handle noisy or incomplete data, and achieve superior results, thereby enhancing the robustness and versatility of shape modeling methodologies.

\section{Conclusion}
This study establishes the effectiveness of \model-based approaches in shape modeling for diverse anatomical datasets, highlighting their strengths in accuracy, adaptability, and uncertainty quantification. The proposed methods, M++Flow and M++AE, demonstrate superior or comparable performance to traditional optimization-based techniques such as ShapeWorks across key metrics, including compactness, generalization, specificity, and distance-based evaluations. Notably, the proposed models achieve high fidelity in representing both simple structures, like the femur, and complex geometries, such as the liver, pancreas, and left atrium, and display reduced computational complexity at inference while maintaining robust performance.

The calibration of aleatoric uncertainty proves to be a significant advantage in evaluating model reliability and identifying outliers. The spatial correlation between uncertainty and prediction errors is particularly evident in datasets with complex or irregular geometries, such as the pancreas and liver. The sample-wise uncertainties are particularly powerful in identifying outliers with more prevalent structural anomalies or segmentation ambiguities. Identifying outliers and quantifying prediction confidence has profound implications for clinical applications, enabling more reliable decision-making and improved model interpretability.

The multi-anatomy modeling experiments further validate the capability of \model~in capturing subtle morphological differences across highly similar structures, as demonstrated by the lumbar vertebrae dataset. The results emphasize the model's precision in distinguishing structural variations. Moreover, the modes of variation analysis align well with known population-level anatomical differences, underscoring the utility of \model~in statistical shape analysis and downstream tasks such as disease localization and group difference analysis.

Overall, integrating deep learning with probabilistic modeling in \model~provides a comprehensive framework for shape modeling. The methods are highly adaptable, efficient, and reliable, making them well-suited for diverse biomedical applications. By addressing challenges like correspondence efficiency, uncertainty quantification, and multi-anatomy modeling, \model~sets a new benchmark for shape analysis tools in clinical and research settings. 
\section{Acknowledgement}
This work was supported by National Institutes of Health under grant numbers NIBIB-U24EB029011, NIAMS-R01AR076120, NHLBI-R01HL135568, and NIBIB- R01EB016701. The content is solely the responsibility of the authors and does not necessarily represent the official views of the National Institutes of Health. The authors thank the ShapeWorks team, the University of Utah Division of Cardiovascular Medicine for providing the left atrium MRI scans and segmentations from the Atrial Fibrillation projects, and the Orthopaedic Research Laboratory at the University of Utah for providing femur CT scans and segmentations.

\bibliographystyle{IEEEtran}
\bibliography{ref}

\begin{thebibliography}{10}
\providecommand{\url}[1]{#1}
\csname url@samestyle\endcsname
\providecommand{\newblock}{\relax}
\providecommand{\bibinfo}[2]{#2}
\providecommand{\BIBentrySTDinterwordspacing}{\spaceskip=0pt\relax}
\providecommand{\BIBentryALTinterwordstretchfactor}{4}
\providecommand{\BIBentryALTinterwordspacing}{\spaceskip=\fontdimen2\font plus
\BIBentryALTinterwordstretchfactor\fontdimen3\font minus \fontdimen4\font\relax}
\providecommand{\BIBforeignlanguage}[2]{{%
\expandafter\ifx\csname l@#1\endcsname\relax
\typeout{** WARNING: IEEEtran.bst: No hyphenation pattern has been}%
\typeout{** loaded for the language `#1'. Using the pattern for}%
\typeout{** the default language instead.}%
\else
\language=\csname l@#1\endcsname
\fi
#2}}
\providecommand{\BIBdecl}{\relax}
\BIBdecl

\bibitem{singh2020evaluation}
B.~Singh, N.~R. Kumar, A.~Balan, M.~Nishan, P.~Haris, M.~Jinisha, and C.~D. Denny, ``Evaluation of normal morphology of mandibular condyle: a radiographic survey,'' \emph{Journal of clinical imaging science}, vol.~10, 2020.

\bibitem{dai2020statistical}
H.~Dai, N.~Pears, W.~Smith, and C.~Duncan, ``Statistical modeling of craniofacial shape and texture,'' \emph{International Journal of Computer Vision}, vol. 128, no.~2, pp. 547--571, 2020.

\bibitem{zhang20243dcmm}
J.~Zhang, K.~Zhou, Y.~Luximon, T.-Y. Lee, and P.~Li, ``3dcmm: 3d comprehensive morphable models with uv-unet for accurate head creation,'' \emph{IEEE Transactions on Multimedia}, 2024.

\bibitem{quiceno2024statistical}
E.~Quiceno, C.~D. Correa, J.~A. Tamayo, and A.~A. Zuleta, ``Statistical models and implant customization in hip arthroplasty: Seeking patient satisfaction through design,'' \emph{Heliyon}, 2024.

\bibitem{khan2022machine}
R.~A. Khan, Y.~Luo, and F.-X. Wu, ``Machine learning based liver disease diagnosis: A systematic review,'' \emph{Neurocomputing}, vol. 468, pp. 492--509, 2022.

\bibitem{schaufelberger2022radiation}
M.~Schaufelberger, R.~K{\"u}hle, A.~Wachter, F.~Weichel, N.~Hagen, F.~Ringwald, U.~Eisenmann, J.~Hoffmann, M.~Engel, C.~Freudlsperger \emph{et~al.}, ``A radiation-free classification pipeline for craniosynostosis using statistical shape modeling,'' \emph{Diagnostics}, vol.~12, no.~7, p. 1516, 2022.

\bibitem{peiffer2022statistical}
M.~Peiffer, A.~Burssens, S.~De~Mits, T.~Heintz, M.~Van~Waeyenberge, K.~Buedts, J.~Victor, and E.~Audenaert, ``Statistical shape model-based tibiofibular assessment of syndesmotic ankle lesions using weight-bearing ct,'' \emph{Journal of Orthopaedic Research{\textregistered}}, vol.~40, no.~12, pp. 2873--2884, 2022.

\bibitem{sophocleous2022feasibility}
F.~Sophocleous, A.~B{\^o}ne, A.~I. Shearn, M.~N.~V. Forte, J.~L. Bruse, M.~Caputo, and G.~Biglino, ``Feasibility of a longitudinal statistical atlas model to study aortic growth in congenital heart disease,'' \emph{Computers in Biology and Medicine}, vol. 144, p. 105326, 2022.

\bibitem{vicory2022statistical}
J.~Vicory, C.~Herz, D.~Allemang, H.~H. Nam, A.~Cianciulli, C.~Vigil, Y.~Han, A.~Lasso, M.~A. Jolley, and B.~Paniagua, ``Statistical shape analysis of the tricuspid valve in hypoplastic left heart syndrome,'' in \emph{Statistical Atlases and Computational Models of the Heart. Multi-Disease, Multi-View, and Multi-Center Right Ventricular Segmentation in Cardiac MRI Challenge: 12th International Workshop, STACOM 2021, Held in Conjunction with MICCAI 2021, Strasbourg, France, September 27, 2021, Revised Selected Papers}.\hskip 1em plus 0.5em minus 0.4em\relax Springer, 2022, pp. 132--140.

\bibitem{merle2019high}
C.~Merle, M.~M. Innmann, W.~Waldstein, E.~C. Pegg, P.~R. Aldinger, H.~S. Gill, D.~W. Murray, and G.~Grammatopoulos, ``High variability of acetabular offset in primary hip osteoarthritis influences acetabular reaming—a computed tomography--based anatomic study,'' \emph{The Journal of Arthroplasty}, vol.~34, no.~8, pp. 1808--1814, 2019.

\bibitem{mulder2022dynamic}
S.~T. Mulder, A.-H. Omidvari, A.~J. Rueten-Budde, P.-H. Huang, K.-H. Kim, B.~Bais, M.~Rousian, R.~Hai, C.~Akgun, J.~R. van Lennep \emph{et~al.}, ``Dynamic digital twin: Diagnosis, treatment, prediction, and prevention of disease during the life course,'' \emph{Journal of Medical Internet Research}, vol.~24, no.~9, p. e35675, 2022.

\bibitem{okegbile2022human}
S.~D. Okegbile, J.~Cai, D.~Niyato, and C.~Yi, ``Human digital twin for personalized healthcare: Vision, architecture and future directions,'' \emph{IEEE network}, vol.~37, no.~2, pp. 262--269, 2022.

\bibitem{durrleman2014morphometry}
S.~Durrleman, M.~Prastawa, N.~Charon, J.~R. Korenberg, S.~Joshi, G.~Gerig, and A.~Trouv{\'e}, ``Morphometry of anatomical shape complexes with dense deformations and sparse parameters,'' \emph{NeuroImage}, vol. 101, pp. 35--49, 2014.

\bibitem{samson2000level}
C.~Samson, L.~Blanc-F{\'e}raud, G.~Aubert, and J.~Zerubia, ``A level set model for image classification,'' \emph{International journal of computer vision}, vol.~40, no.~3, pp. 187--197, 2000.

\bibitem{styner2006framework}
M.~Styner, I.~Oguz, S.~Xu, C.~Brechb{\"u}hler, D.~Pantazis, J.~J. Levitt, M.~E. Shenton, and G.~Gerig, ``Framework for the statistical shape analysis of brain structures using spharm-pdm,'' \emph{The insight journal}, no. 1071, p. 242, 2006.

\bibitem{durrleman2009statistical}
S.~Durrleman, X.~Pennec, A.~Trouv{\'e}, and N.~Ayache, ``Statistical models of sets of curves and surfaces based on currents,'' \emph{Medical image analysis}, vol.~13, no.~5, pp. 793--808, 2009.

\bibitem{cates2007shape}
J.~Cates, P.~T. Fletcher, M.~Styner, M.~Shenton, and R.~Whitaker, ``Shape modeling and analysis with entropy-based particle systems,'' in \emph{Information Processing in Medical Imaging: 20th International Conference, IPMI 2007, Kerkrade, The Netherlands, July 2-6, 2007. Proceedings 20}.\hskip 1em plus 0.5em minus 0.4em\relax Springer, 2007, pp. 333--345.

\bibitem{cates2008particle}
J.~Cates, P.~T. Fletcher, M.~Styner, H.~C. Hazlett, and R.~Whitaker, ``Particle-based shape analysis of multi-object complexes,'' in \emph{International Conference on Medical Image Computing and Computer-Assisted Intervention}.\hskip 1em plus 0.5em minus 0.4em\relax Springer, 2008, pp. 477--485.

\bibitem{cerrolaza2019computational}
J.~J. Cerrolaza, M.~L. Picazo, L.~Humbert, Y.~Sato, D.~Rueckert, M.~{\'A}.~G. Ballester, and M.~G. Linguraru, ``Computational anatomy for multi-organ analysis in medical imaging: A review,'' \emph{Medical Image Analysis}, vol.~56, pp. 44--67, 2019.

\bibitem{iyer2023mesh2ssm}
K.~Iyer and S.~Y. Elhabian, ``Mesh2ssm: From surface meshes to statistical shape models of anatomy,'' in \emph{International Conference on Medical Image Computing and Computer-Assisted Intervention}.\hskip 1em plus 0.5em minus 0.4em\relax Springer, 2023, pp. 615--625.

\bibitem{adams2023point2ssm}
J.~Adams and S.~Elhabian, ``Point2ssm: Learning morphological variations of anatomies from point cloud,'' \emph{arXiv preprint arXiv:2305.14486}, 2023.

\bibitem{ludke2022landmark}
D.~L{\"u}dke, T.~Amiranashvili, F.~Ambellan, I.~Ezhov, B.~H. Menze, and S.~Zachow, ``Landmark-free statistical shape modeling via neural flow deformations,'' in \emph{Medical Image Computing and Computer Assisted Intervention--MICCAI 2022: 25th International Conference, Singapore, September 18--22, 2022, Proceedings, Part II}.\hskip 1em plus 0.5em minus 0.4em\relax Springer, 2022, pp. 453--463.

\bibitem{bhalodia2024deepssm}
R.~Bhalodia, S.~Elhabian, J.~Adams, W.~Tao, L.~Kavan, and R.~Whitaker, ``Deepssm: A blueprint for image-to-shape deep learning models,'' \emph{Medical Image Analysis}, vol.~91, p. 103034, 2024.

\bibitem{bhalodia2018deepssm}
R.~Bhalodia, S.~Y. Elhabian, L.~Kavan, and R.~T. Whitaker, ``Deepssm: a deep learning framework for statistical shape modeling from raw images,'' in \emph{Shape in Medical Imaging: International Workshop, ShapeMI 2018, Held in Conjunction with MICCAI 2018, Granada, Spain, September 20, 2018, Proceedings}.\hskip 1em plus 0.5em minus 0.4em\relax Springer, 2018, pp. 244--257.

\bibitem{el2024universal}
N.~El~Amrani, D.~Cao, and F.~Bernard, ``A universal and flexible framework for unsupervised statistical shape model learning,'' in \emph{International Conference on Medical Image Computing and Computer-Assisted Intervention}.\hskip 1em plus 0.5em minus 0.4em\relax Springer, 2024, pp. 26--36.

\bibitem{kalaie2024end}
S.~Kalaie, A.~Bulpitt, A.~F. Frangi, and A.~Gooya, ``An end-to-end deep learning generative framework for refinable shape matching and generation,'' \emph{arXiv preprint arXiv:2403.06317}, 2024.

\bibitem{kingma2019introduction}
D.~P. Kingma, M.~Welling \emph{et~al.}, ``An introduction to variational autoencoders,'' \emph{Foundations and Trends{\textregistered} in Machine Learning}, vol.~12, no.~4, pp. 307--392, 2019.

\bibitem{rezende2015variational}
D.~Rezende and S.~Mohamed, ``Variational inference with normalizing flows,'' in \emph{International conference on machine learning}.\hskip 1em plus 0.5em minus 0.4em\relax PMLR, 2015, pp. 1530--1538.

\bibitem{dinh2016density}
L.~Dinh, J.~Sohl-Dickstein, and S.~Bengio, ``Density estimation using real nvp,'' \emph{arXiv preprint arXiv:1605.08803}, 2016.

\bibitem{paulsen2002building}
R.~Paulsen, R.~Larsen, C.~Nielsen, S.~Laugesen, and B.~Ersb{\o}ll, ``Building and testing a statistical shape model of the human ear canal,'' in \emph{International Conference on Medical Image Computing and Computer-Assisted Intervention}.\hskip 1em plus 0.5em minus 0.4em\relax Springer, 2002, pp. 373--380.

\bibitem{heitz2005statistical}
G.~Heitz, T.~Rohlfing, and C.~R. Maurer~Jr, ``Statistical shape model generation using nonrigid deformation of a template mesh,'' in \emph{Medical Imaging 2005: Image Processing}, vol. 5747.\hskip 1em plus 0.5em minus 0.4em\relax SPIE, 2005, pp. 1411--1421.

\bibitem{mcinerney1996deformable}
T.~McInerney and D.~Terzopoulos, ``Deformable models in medical image analysis,'' in \emph{Proceedings of the workshop on mathematical methods in biomedical image analysis}.\hskip 1em plus 0.5em minus 0.4em\relax IEEE, 1996, pp. 171--180.

\bibitem{cates2017shapeworks}
J.~Cates, S.~Elhabian, and R.~Whitaker, ``Shapeworks: Particle-based shape correspondence and visualization software,'' in \emph{Statistical Shape and Deformation Analysis}.\hskip 1em plus 0.5em minus 0.4em\relax Elsevier, 2017, pp. 257--298.

\bibitem{oguz2016entropy}
I.~Oguz, J.~Cates, M.~Datar, B.~Paniagua, T.~Fletcher, C.~Vachet, M.~Styner, and R.~Whitaker, ``Entropy-based particle correspondence for shape populations,'' \emph{International journal of computer assisted radiology and surgery}, vol.~11, pp. 1221--1232, 2016.

\bibitem{adams2022images}
J.~Adams and S.~Elhabian, ``From images to probabilistic anatomical shapes: A deep variational bottleneck approach,'' in \emph{Medical Image Computing and Computer Assisted Intervention--MICCAI 2022: 25th International Conference, Singapore, September 18--22, 2022, Proceedings, Part II}.\hskip 1em plus 0.5em minus 0.4em\relax Springer, 2022, pp. 474--484.

\bibitem{lang2021dpc}
I.~Lang, D.~Ginzburg, S.~Avidan, and D.~Raviv, ``Dpc: Unsupervised deep point correspondence via cross and self construction,'' in \emph{2021 International Conference on 3D Vision (3DV)}.\hskip 1em plus 0.5em minus 0.4em\relax IEEE, 2021, pp. 1442--1451.

\bibitem{wang2019dynamic}
Y.~Wang, Y.~Sun, Z.~Liu, S.~E. Sarma, M.~M. Bronstein, and J.~M. Solomon, ``Dynamic graph cnn for learning on point clouds,'' \emph{ACM Transactions on Graphics (tog)}, vol.~38, no.~5, pp. 1--12, 2019.

\bibitem{chen2020unsupervised}
N.~Chen, L.~Liu, Z.~Cui, R.~Chen, D.~Ceylan, C.~Tu, and W.~Wang, ``Unsupervised learning of intrinsic structural representation points,'' in \emph{Proceedings of the IEEE/CVF conference on computer vision and pattern recognition}, 2020, pp. 9121--9130.

\bibitem{chang2015shapenet}
A.~X. Chang, T.~Funkhouser, L.~Guibas, P.~Hanrahan, Q.~Huang, Z.~Li, S.~Savarese, M.~Savva, S.~Song, H.~Su \emph{et~al.}, ``Shapenet: An information-rich 3d model repository,'' \emph{arXiv preprint arXiv:1512.03012}, 2015.

\bibitem{choy20163d}
C.~B. Choy, D.~Xu, J.~Gwak, K.~Chen, and S.~Savarese, ``3d-r2n2: A unified approach for single and multi-view 3d object reconstruction,'' in \emph{Computer Vision--ECCV 2016: 14th European Conference, Amsterdam, The Netherlands, October 11-14, 2016, Proceedings, Part VIII 14}.\hskip 1em plus 0.5em minus 0.4em\relax Springer, 2016, pp. 628--644.

\bibitem{zhang2019graph}
S.~Zhang, H.~Tong, J.~Xu, and R.~Maciejewski, ``Graph convolutional networks: a comprehensive review,'' \emph{Computational Social Networks}, vol.~6, no.~1, pp. 1--23, 2019.

\bibitem{kipf2016semi}
T.~N. Kipf and M.~Welling, ``Semi-supervised classification with graph convolutional networks,'' \emph{arXiv preprint arXiv:1609.02907}, 2016.

\bibitem{hanocka2019meshcnn}
R.~Hanocka, A.~Hertz, N.~Fish, R.~Giryes, S.~Fleishman, and D.~Cohen-Or, ``Meshcnn: a network with an edge,'' \emph{ACM Transactions on Graphics (ToG)}, vol.~38, no.~4, pp. 1--12, 2019.

\bibitem{defferrard2016convolutional}
M.~Defferrard, X.~Bresson, and P.~Vandergheynst, ``Convolutional neural networks on graphs with fast localized spectral filtering,'' \emph{Advances in neural information processing systems}, vol.~29, 2016.

\bibitem{jiang2020shapeflow}
C.~Jiang, J.~Huang, A.~Tagliasacchi, and L.~J. Guibas, ``Shapeflow: Learnable deformation flows among 3d shapes,'' \emph{Advances in Neural Information Processing Systems}, vol.~33, pp. 9745--9757, 2020.

\bibitem{chen2019net}
Z.~Chen, ``Im-net: Learning implicit fields for generative shape modeling,'' 2019.

\bibitem{alemi2018fixing}
A.~Alemi, B.~Poole, I.~Fischer, J.~Dillon, R.~A. Saurous, and K.~Murphy, ``Fixing a broken elbo,'' in \emph{International conference on machine learning}.\hskip 1em plus 0.5em minus 0.4em\relax PMLR, 2018, pp. 159--168.

\bibitem{chen2023learning}
S.~Chen, S.~Ding, Y.~Karayiannidis, and M.~Bj{\"o}rkman, ``Learning continuous normalizing flows for faster convergence to target distribution via ascent regularizations,'' in \emph{The Eleventh International Conference on Learning Representations}, 2023.

\bibitem{bhalodia2020dpvaes}
R.~Bhalodia, I.~Lee, and S.~Elhabian, ``dpvaes: Fixing sample generation for regularized vaes,'' in \emph{Proceedings of the Asian Conference on Computer Vision}, 2020.

\bibitem{simpson2019large}
A.~L. Simpson, M.~Antonelli, S.~Bakas, M.~Bilello, K.~Farahani, B.~Van~Ginneken, A.~Kopp-Schneider, B.~A. Landman, G.~Litjens, B.~Menze \emph{et~al.}, ``A large annotated medical image dataset for the development and evaluation of segmentation algorithms,'' \emph{arXiv preprint arXiv:1902.09063}, 2019.

\bibitem{Ma-2021-AbdomenCT-1K}
J.~Ma, Y.~Zhang, S.~Gu, C.~Zhu, C.~Ge, Y.~Zhang, X.~An, C.~Wang, Q.~Wang, X.~Liu, S.~Cao, Q.~Zhang, S.~Liu, Y.~Wang, Y.~Li, J.~He, and X.~Yang, ``Abdomenct-1k: Is abdominal organ segmentation a solved problem?'' \emph{IEEE Transactions on Pattern Analysis and Machine Intelligence}, vol.~44, no.~10, pp. 6695--6714, 2022.

\bibitem{bautz2023unsupervised}
L.~Bautz, ``Unsupervised shape correspondence estimation for anatomical shapes,'' Master's thesis, 2023.

\bibitem{casaclang2008structural}
G.~Casaclang-Verzosa, B.~J. Gersh, and T.~S. Tsang, ``Structural and functional remodeling of the left atrium: clinical and therapeutic implications for atrial fibrillation,'' \emph{Journal of the American College of Cardiology}, vol.~51, no.~1, pp. 1--11, 2008.

\bibitem{higaki2023liver}
A.~Higaki, A.~Kanki, A.~Yamamoto, Y.~Ueda, K.~Moriya, H.~Sanai, H.~Sotozono, and T.~Tamada, ``Liver cirrhosis: relationship between fibrosis-associated hepatic morphological changes and portal hemodynamics using four-dimensional flow magnetic resonance imaging,'' \emph{Japanese Journal of Radiology}, vol.~41, no.~6, pp. 625--636, 2023.

\bibitem{sekuboyina2021verse}
A.~Sekuboyina, M.~E. Husseini, A.~Bayat, M.~L{\"o}ffler, H.~Liebl, H.~Li, G.~Tetteh, J.~Kuka{\v{c}}ka, C.~Payer, D.~{\v{S}}tern \emph{et~al.}, ``Verse: a vertebrae labelling and segmentation benchmark for multi-detector ct images,'' \emph{Medical image analysis}, vol.~73, p. 102166, 2021.

\bibitem{araslanovdiffcd}
N.~Araslanov and D.~Cremers, ``Diffcd: A symmetric differentiable chamfer distance for neural implicit surface fitting.''

\bibitem{wang2023neural}
Z.~Wang, P.~Wang, P.~Wang, Q.~Dong, J.~Gao, S.~Chen, S.~Xin, C.~Tu, and W.~Wang, ``Neural-imls: Self-supervised implicit moving least-squares network for surface reconstruction,'' \emph{IEEE Transactions on Visualization and Computer Graphics}, 2023.

\end{thebibliography}
\vskip -2.5\baselineskip plus -1fil
\begin{IEEEbiographynophoto}{Krithika Iyer} is a computing PhD candidate at the Kahlert School of Computing and 
the Scientific Computing and Imaging Institute at the University of Utah in the Image Analysis track. She received her B.E. in Electronics \& Telecommunication in 2015. Her research interests include medical image analysis, probabilistic modeling, deep learning, and statistical shape modeling.\end{IEEEbiographynophoto}
\vskip -2.5\baselineskip plus -1fil
\begin{IEEEbiographynophoto}{Mokshagna Sai Teja Karanam} is a computing PhD student at the Kahlert School of Computing and the Scientific Computing and Imaging Institute at the University of Utah in the Image Analysis track. He received his B.E. in Computer Science in 2020 and his Masters in 2024. His research interests include deep learning, computer vision and statistical shape modeling.\end{IEEEbiographynophoto}
\vskip -2.5\baselineskip plus -1fil
\begin{IEEEbiographynophoto}{Shireen Elhabian} is a faculty member at the Kahlert School of Computing and the Scientific Computing and Imaging Institute at the University of Utah. Her research focuses on enhancing diagnostic accuracy through deep learning, probabilistic modeling, and advanced computer vision algorithms. She has published over 100 peer-reviewed publications in prestigious journals and conferences, including IEEE-TMI, MedIA, ICLR, CVPR, ICCV, MICCAI, IPMI, AAAI. \end{IEEEbiographynophoto}
\vskip -3\baselineskip plus -1fil
\appendices

\section{Modes of Variation}
\begin{figure}
    \centering
    \includegraphics[width=\linewidth]{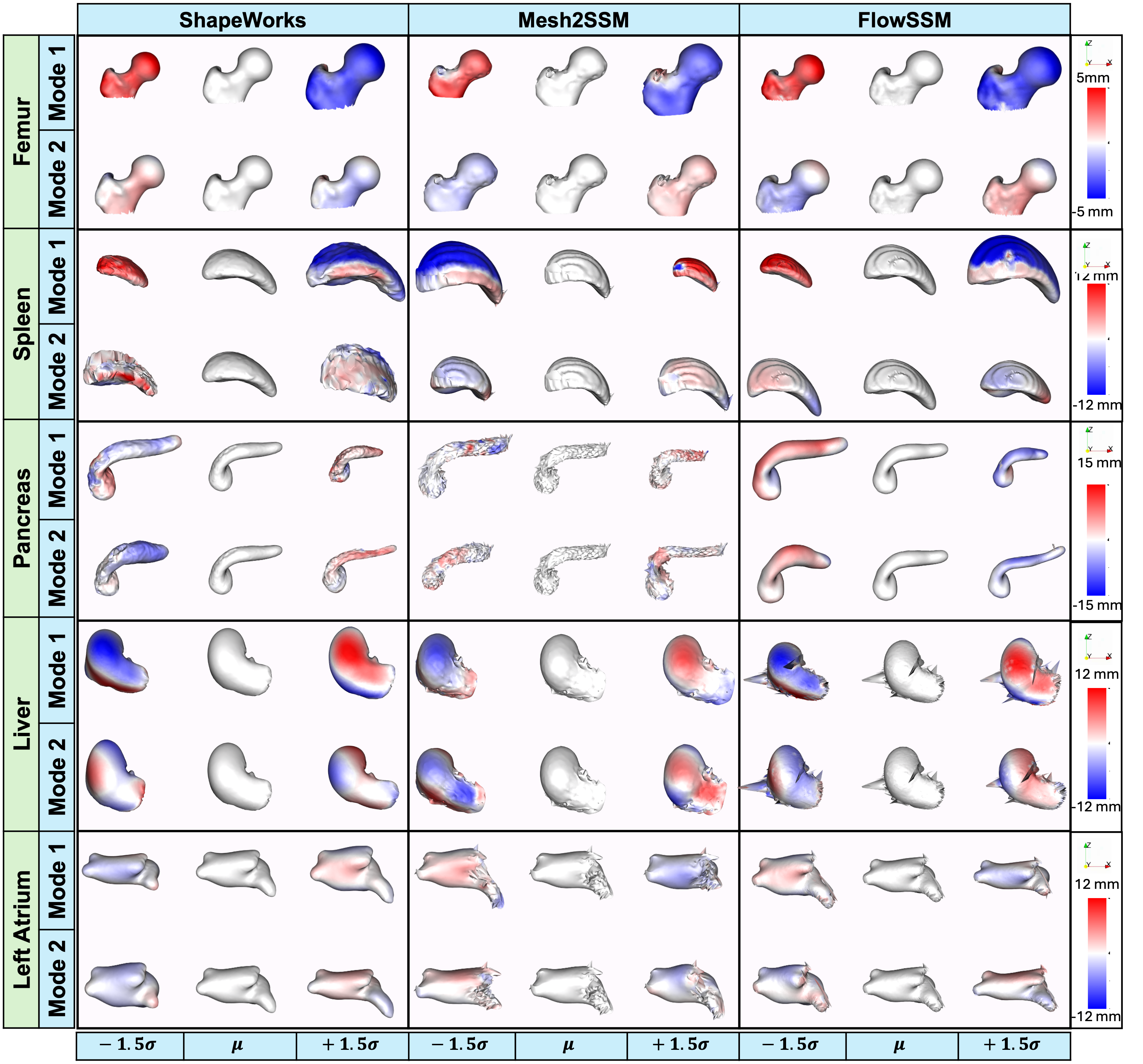}
    \caption{\textbf{Modes of Variation:} The first two modes were identified by performing PCA on the predicted correspondences.}
    \label{fig:enter-label}
\end{figure}

\end{document}